\documentclass[11pt]{article}

\usepackage[final]{acl}
\usepackage[table]{xcolor}
\usepackage{multirow}
\usepackage{graphicx}   
\usepackage{times}
\usepackage{latexsym}
\usepackage{amsmath} 

\usepackage[T1]{fontenc}
\usepackage{multirow}
\usepackage{xcolor}

\usepackage[utf8]{inputenc}

\usepackage{microtype}
\usepackage{amsmath}      
\usepackage{amssymb}      
\usepackage{algorithm}    
\usepackage{algpseudocode} 
\usepackage{inconsolata}
\usepackage{booktabs} 
\usepackage{graphicx}

\title{Bring Back Language to Actions: Restoring VLM capabilities in VLA Models via Parameter Merging}

\author{
 \textbf{Mohamed Salim Aissi\textsuperscript{1}},
  \textbf{Olivier Sigaud\textsuperscript{1}},
  \textbf{Laure Soulier \textsuperscript{1}},
  \textbf{Nicolas Thome \textsuperscript{1,2,3}}
\\
  \textsuperscript{1}Sorbonne Université, CNRS, ISIR, F-75005 Paris, France \\
  \textsuperscript{2}Institut universitaire de France (IUF) \\
  \textsuperscript{3}{International Laboratory on Learning Systems (ILLS)}
  \\ \{salim.aissi, laure.soulier, olivier.sigaud, nicolas.thome\}@isir.upmc.fr
\\ 
}

\begin{document}
\maketitle
\begin{figure*}[ht]
    \centering
    \includegraphics[width=0.52\linewidth]{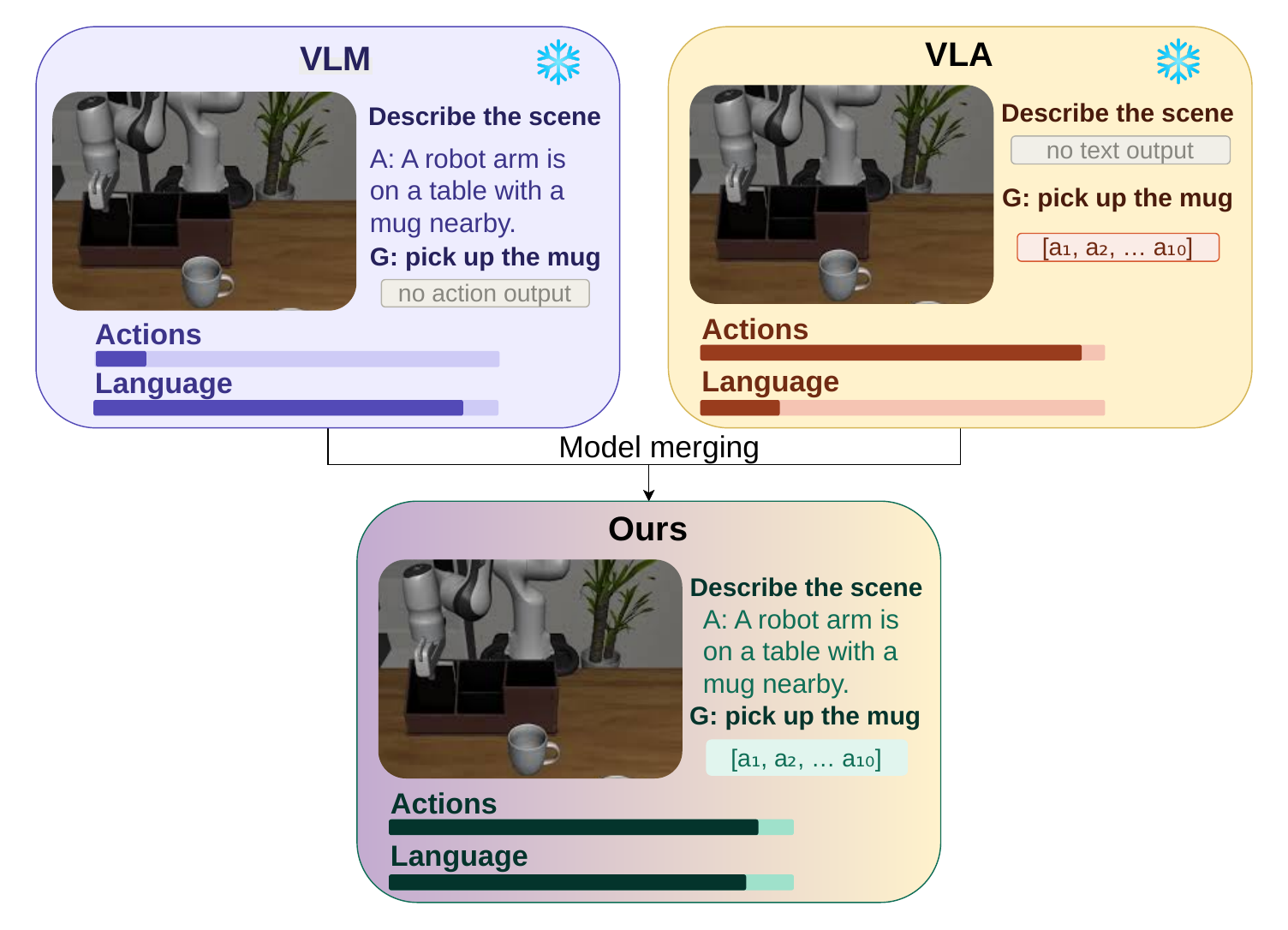}\includegraphics[width=0.47\linewidth]{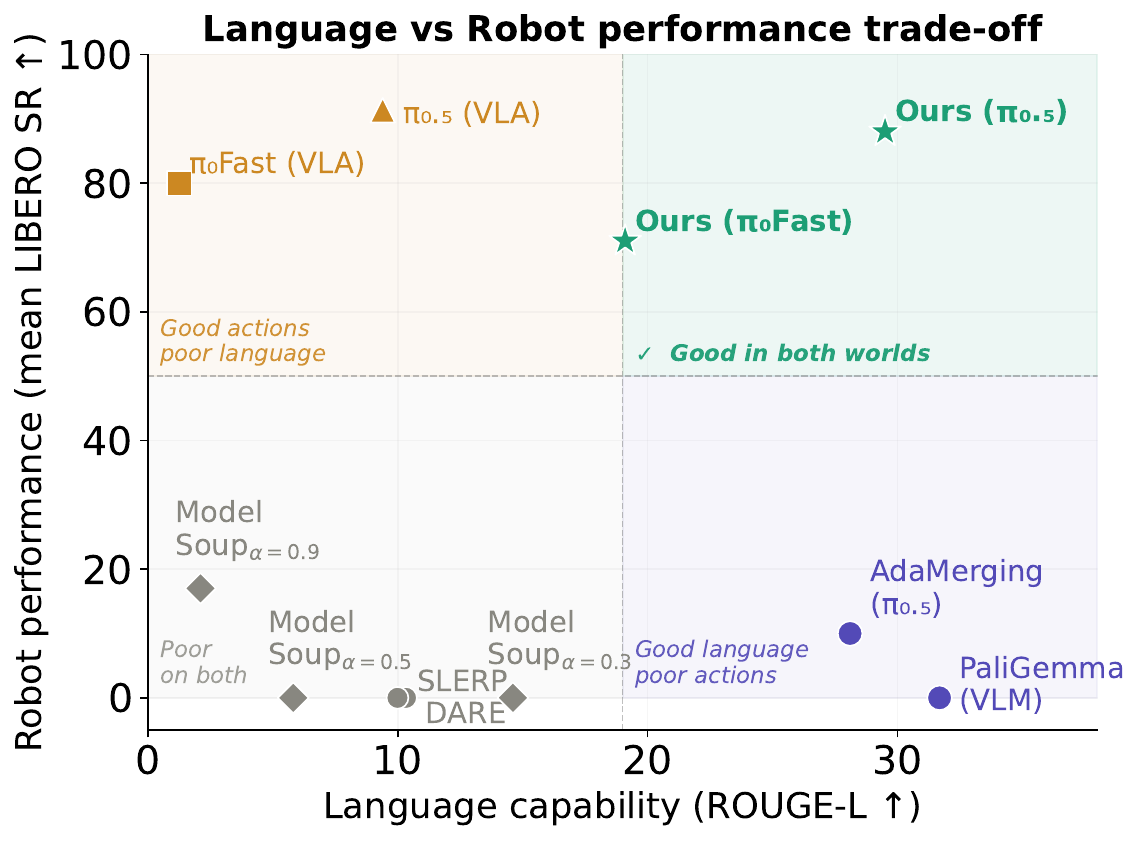}
    \caption{\textit{(Left)} Given the same visual scene, a VLM responds accurately to descriptive language prompts but produces no action output. Fine-tuning into a VLA enables action prediction from a task goal, but at the cost of language grounding — the VLA generates incoherent text when prompted with the same descriptive query. Our merged model recovers both capabilities, producing coherent language responses while retaining the ability to predict actions.  \textit{(Right)} The VLM achieves strong language performance but cannot act, while VLAs excel at manipulation but lose language capability. Our merged models are the only ones that achieve the best of both worlds, reaching the upper-right region of the trade-off space.}
        \vspace{-0.4cm}
    \label{fig:introfig}
\end{figure*}

\begin{abstract}
Vision-Language-Action models (VLAs) are obtained by fine-tuning pre-trained Vision-Language Models (VLMs) on robot demonstration data, enabling action prediction from visual observations and language instructions. However, this fine-tuning induces catastrophic forgetting, substantially degrading the semantic and visual grounding capabilities of the original VLM. We address this by merging the parameters of a VLM and its VLA derivative without retraining either model. Concretely, we extract the task vector — the difference between VLA and VLM weights — and decompose it via Singular Value Decomposition. For each component, we learn a scalar mixing coefficient controlling how much of the VLA adaptation is retained, optimised to jointly preserve action prediction and vision-language capabilities. 
Experiments show that our merged model performs well on both axes simultaneously, consistently outperforming established merging baselines.
\end{abstract}

\section{Introduction}

The ability to reason about complex environments and to plan sequences of actions has long been a central challenge in artificial intelligence. Large Language Models (LLMs) and Vision-Language Models (VLMs) have made significant strides in this direction, combining reasoning, visual perception, and scene understanding into unified models capable of high-level instruction following, task decomposition, and environment description \citep{wei2023chainofthoughtpromptingelicitsreasoning,yao2023reactsynergizingreasoningacting,radford2021learningtransferablevisualmodels,liu2023visualinstructiontuning,dai2026languagemovementprimitivesgrounding,aissi2026prismperceptionreasoninginterleaved} — capabilities that make them compelling building blocks for autonomous robotic agents. However, these models remain confined to perception and reasoning, and cannot directly produce the low-level motor commands required for physical interaction.

Vision-Language-Action models (VLAs) bridge this gap by fine-tuning pre-trained VLMs on robot demonstration data \citep{embodimentcollaboration2025openxembodimentroboticlearning,walke2023bridgedata}, enabling them to predict actions directly from visual observations and language instructions \cite{brohan2023rt2visionlanguageactionmodelstransfer,kim2024openvlaopensourcevisionlanguageactionmodel}. However, this fine-tuning comes at a significant cost: the action objective progressively overwrites the weight-space directions encoding visual-language understanding \cite{hancock2025actionslanguagefinetuningvlms}, yielding a specialised model that excels at low-level control but can no longer reliably perform the high-level semantic reasoning needed for robust, generalist robotic agents.

This tension between action specialisation and semantic preservation is a fundamental obstacle to building robotic systems that are both capable of language modelling and fine-grained motor control (see Figure~\ref{fig:introfig}). Existing approaches to mitigating this degradation typically require architectural modifications or costly co-training on mixed datasets \cite{Zawalski24-ecot,hancock2025actionslanguagefinetuningvlms}, making them difficult to apply post-hoc to already fine-tuned VLAs. Model merging offers a compelling alternative \citep{BECAME,yadav2025robust}: by combining the parameters of a VLM and a VLA, one can in principle recover a model that retains both action prediction and vision-language capabilities. However, naive uniform blending treats all parameters equally and fails to account for the heterogeneous roles played by different weight matrices across the two models, yielding suboptimal trade-offs between semantic and action performance.

In this work \footnote{The code can be found
\href{https://github.com/sal1717lim/bring-back-language-to-actions}{here}}, we investigate how to merge a VLM and a VLA such that the resulting model retains the action capabilities acquired during fine-tuning while recovering the vision-language understanding of the original VLM. To this end, we extract the task vector by subtracting the VLM weights from those of the VLA and decompose it using Singular Value Decomposition to expose the individual components of this adaptation. For each such direction, we learn a scalar coefficient that controls how much of the VLA adaptation is retained in the merged model, jointly preserving action prediction performance and the original vision-language capabilities of the VLM. We demonstrate that this approach consistently outperforms uniform blending and established merging baselines across both vision-language benchmarks and robotic manipulation tasks, on two distinct VLA architectures.
 
\noindent Our contributions are as follows:

\noindent $\bullet$ \textbf{A post-hoc, lightweight merging method} that decomposes the VLM--VLA task vector via SVD and learns per-component coefficients to jointly recover language understanding and preserve action capabilities, requiring only the optimisation of a small set of scalar merging parameters while keeping both source models fully frozen.

\noindent $\bullet$   \textbf{An ablation study justifying fine-grained merging}, showing per-component coefficients are necessary where global or per-layer scalars fail.

\noindent $\bullet$  \textbf{A structural analysis of the fine-tuning directions} at multiple scales of the SVD decomposition, identifying which components of the adaptation must be preserved to reconcile vision-language understanding and action prediction in the merged model.

Together, these contributions offer a practical and substantially cheaper path toward robotic agents that combine reliable low-level control with the high-level semantic reasoning of their underlying VLM.
\section{Related Work}
\subsection{Vision Language Action Models}
Vision-Language-Action (VLA) models extend Vision-Language Models (VLMs) by fine‑tuning them on large‑scale robot demonstration data \citep{embodimentcollaboration2025openxembodimentroboticlearning}, enabling them to act as generalist policies that directly generate low‑level actions \citep{brohan2023rt2visionlanguageactionmodelstransfer,octomodelteam2024octoopensourcegeneralistrobot,kim2024openvlaopensourcevisionlanguageactionmodel}. Actions are produced either autoregressively \citep{kim2024openvlaopensourcevisionlanguageactionmodel,pertsch2025fastefficientactiontokenization} or via continuous decoding such as flow matching \citep{black2026pi0visionlanguageactionflowmodel}. However, this fine-tuning causes them to largely lose their original vision–language capabilities through catastrophic forgetting.

Some works sidestep this issue by keeping the VLM frozen and training only a lightweight action module \citep{shukor2025smolvlavisionlanguageactionmodelaffordable,wen2025tinyvlafastdataefficientvisionlanguageaction}, but this limits the model's ability to fully adapt its representations for embodied control. Others reintroduce text tasks during training either by training VLAs to generate both actions and text simultaneously \citep{Zawalski24-ecot,intelligence2025pi05visionlanguageactionmodelopenworld} or by aligning action representations with the language objective \citep{hancock2025actionslanguagefinetuningvlms}. Both approaches require large-scale co-training to recover knowledge already present in the initial VLM. In contrast, our approach operates post-hoc on an already fine-tuned VLA, requiring no additional training of either source model.

\begin{figure*}[ht]
    \centering
    \includegraphics[width=0.88\linewidth]{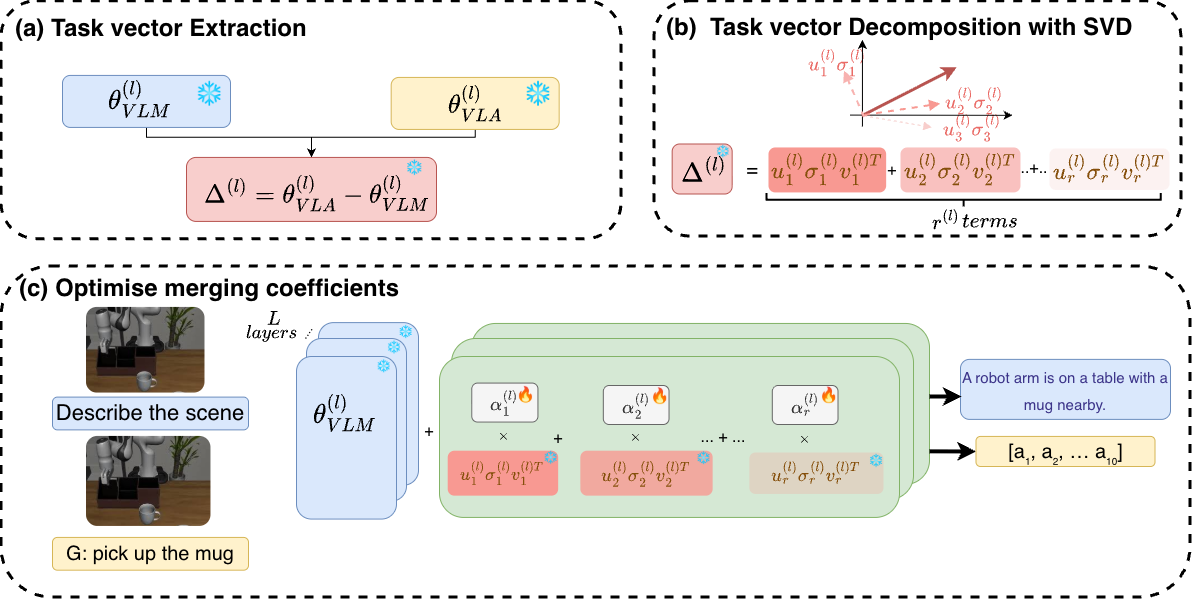}
    \caption{Overview of our VLM--VLA merging method. \textbf{(a)} The task vector $\Delta^{(l)}$ is extracted by subtracting the VLM weights from the VLA weights for each layer $l$. \textbf{(b)} $\Delta^{(l)}$ is decomposed via Singular Value Decomposition, revealing the orthogonal directions of adaptation introduced during VLA fine-tuning. \textbf{(c)} A scalar mixing coefficient $\boldsymbol{\alpha_k^{(l)}}$ is assigned to each singular component, controlling how much of each direction is retained in the merged model. The coefficients $\boldsymbol{\alpha}$ are optimised by minimising a composite objective that jointly penalises degradation of language generation quality and action prediction accuracy with respect to both source models.}
    \label{fig:placeholder}
    \vspace{-0.4cm}
\end{figure*}

\subsection{Model Merging}
Model merging has emerged as a cost-effective alternative to full retraining, enabling the combination of specialised models into a single capable model without additional data or compute. \textbf{Model Soup}~\cite{wortsman2022modelsoupsaveragingweights} averages the weights of models fine-tuned from the same checkpoint. \textbf{SLERP} \citep{jang2024sphericallinearinterpolationtextanchoring} interpolates along the spherical geodesic rather than linearly, better preserving parameter norms. \textbf{TIES}~\cite{yadav2023tiesmergingresolvinginterferencemerging} resolves sign conflicts across multiple task vectors before averaging, requiring at least two independently fine-tuned models and thus not directly applicable to our setting. \textbf{DARE}~\cite{yu2024languagemodelssupermario} randomly drops a fraction of task vector parameters before rescaling the remainder. \textbf{Fisher Merging}~\cite{matena2022mergingmodelsfisherweightedaveraging} weights each parameter by its diagonal Fisher information, but requires computing and storing a matrix as large as the full model, making it costly at billion-parameter scale. \textbf{AdaMerging}~\cite{yang2024adamergingadaptivemodelmerging} learns task-wise or layer-wise merging weights but does not decompose the task vector into finer-grained components.
All these methods operate with fixed or coarse-grained coefficients and none have been applied to reconciling a VLM with its VLA derivative. Closer to our work, decomposing task vectors via SVD has already been proposed to reduce interference between models trained on different tasks \citet{Gargiulo_2025_CVPR}. We build on this insight and extend it to the VLM--VLA setting, where the objective is to reconcile two fundamentally different capabilities — language understanding and action prediction — rather than reduce task interference.
\section{Method}
\label{sec:method}

\subsection{Problem Formulation}

Let $\theta_{\text{VLM}}$ and $\theta_{\text{VLA}}$ denote the parameters of a pre-trained Vision-Language Model and a fine-tuned Vision-Language-Action model, respectively. Both models share a common backbone architecture. The VLM initially possesses strong language and visual grounding capabilities, while the VLA has been further trained to predict robotic actions, albeit at the cost of partial degradation of its original vision-language abilities. Our merging procedure is applied to all layers that are common to both architectures; VLA-specific components such as the action head are left unchanged and carried over directly into the merged model.

Our goal is to construct a merged model $\theta_{\text{merged}}$ that jointly retains the action prediction capabilities of the VLA and the vision-language understanding of the VLM (Figure~\ref{fig:placeholder}a). We define it as a weighted combination of both source models, controlled by a set of mixing coefficients $\boldsymbol{\alpha}$:
\begin{equation}
\small
    \theta_{\text{merged}}^{(l)} = (1 - \boldsymbol{\alpha^{(l)}})\,\theta_{\text{VLM}}^{(l)} + \boldsymbol{\alpha^{(l)}}\,\theta_{\text{VLA}}^{(l)},    l \in \{1, \dots, L\}
    \label{eq:linear_merge}
\end{equation}
where $L$ is the total number of layers and $\alpha^{(l)} \in [0, 1]$ controls the contribution of each source model at layer $l$. The central question is how to determine $\boldsymbol{\alpha}$ — and at what granularity — such that the merged model best preserves both capabilities. In the following, we show that operating at the level of individual singular components of the per-layer task vector provides the necessary fine-grained control.

\subsection{Task Vector and SVD Decomposition}

For each layer $l$, we define the \textit{task vector} as the difference between the VLA and VLM weight matrices (Figure~\ref{fig:placeholder}a):
\begin{equation}
    \Delta^{(l)} = \theta_{\text{VLA}}^{(l)} - \theta_{\text{VLM}}^{(l)}
    \label{eq:task_vector}
\end{equation}

Substituting into Eq.~\eqref{eq:linear_merge}, the merged model can be equivalently written as:
\begin{equation}
    \theta_{\text{merged}}^{(l)} = \theta_{\text{VLM}}^{(l)} + \boldsymbol{\alpha^{(l)}}\,\Delta^{(l)}
    \label{eq:task_merge}
\end{equation}

This formulation shows that merging amounts to scaling the task vector by $\boldsymbol{\alpha^{(l)}}$ at each layer. To obtain finer-grained control, we decompose $\Delta^{(l)}$ into its individual directions via Singular Value Decomposition, as \cite{Gargiulo_2025_CVPR} (Figure~\ref{fig:placeholder}b), allowing a separate coefficient to be assigned to each direction independently:
\begin{equation}
    \Delta^{(l)} = \sum_{k=1}^{r^{(l)}} \sigma_k^{(l)}\, \mathbf{u}_k^{(l)} \mathbf{v}_k^{(l)\top}
    \label{eq:svd}
\end{equation}
where $\sigma_1^{(l)} \geq \dots \geq \sigma_{r^{(l)}}^{(l)}$ are the singular values ranked by importance and $r^{(l)}$ is the rank of $\Delta^{(l)}$. Each term $\sigma_k^{(l)} \mathbf{u}_k^{(l)} \mathbf{v}_k^{(l)\top}$ captures a distinct direction of adaptation introduced during VLA fine-tuning.

Substituting the SVD decomposition of Eq.~\eqref{eq:svd} into Eq.~\eqref{eq:task_merge}, the merged model becomes:
\begin{equation}
    \theta_{\text{merged}}^{(l)} = \theta_{\text{VLM}}^{(l)} + \boldsymbol{\alpha^{(l)}} \sum_{k=1}^{r^{(l)}} \sigma_k^{(l)}\, \mathbf{u}_k^{(l)} \mathbf{v}_k^{(l)\top}
    \label{eq:merge_svd}
\end{equation}

We now replace the single scalar $\boldsymbol{\alpha^{(l)}}$ by a per-component coefficient $\boldsymbol{\alpha_k^{(l)}} \in [0,1]$ (Figure~\ref{fig:placeholder}c), one for each singular direction, yielding our final merging equation:
\begin{equation}
    \theta_{\text{merged}}^{(l)} = \theta_{\text{VLM Prism}}^{(l)} + \sum_{k=1}^{r^{(l)}} \boldsymbol{\alpha_k^{(l)}}\, \sigma_k^{(l)}\, \mathbf{u}_k^{(l)} \mathbf{v}_k^{(l)\top}
    \label{eq:merge}
\end{equation}

When all $\boldsymbol{\alpha_k^{(l)}} = 0$, the layer reduces to the VLM. When all $\boldsymbol{\alpha_k^{(l)}} = 1$, it recovers the full VLA adaptation. The global scalar merging of Eq.~\eqref{eq:task_merge} is recovered as the special case where all coefficients are equal to $\alpha$. Importantly, both $\theta_{\text{VLM}}$ and the SVD decomposition of the task vector are frozen throughout — the only free parameters are the mixing coefficients $\boldsymbol{\alpha}$, which are optimised to simultaneously satisfy the language understanding objective of the VLM and the action prediction objective of the VLA, as described in Section~\ref{sec:fitness}.


\subsection{Training Objective}
\label{sec:fitness}
 
To optimise the mixing coefficients $\boldsymbol{\alpha}$, we define a composite objective that captures both language fidelity and action prediction accuracy, while penalising excessive deviation from either source model:

{\vspace{-0.5cm}
\begin{equation}
\begin{split}
    \mathcal{L}(\boldsymbol{\alpha}) =\;
    &\lambda_1\mathcal{L}_{\text{text}}(\boldsymbol{\alpha})
    + \lambda_2 \cdot D_{\text{KL}}\!\left(\pi_{\boldsymbol{\alpha}} \;\|\; \pi_{\text{VLM}}\right) \\
    &+ \lambda_3\mathcal{L}_{\text{act}}(\boldsymbol{\alpha})
    + \lambda_4 \cdot D_{\text{KL}}\!\left(\pi_{\boldsymbol{\alpha}} \;\|\; \pi_{\text{VLA}}\right)
\end{split}
\label{eq:objective}
\end{equation}
}

where $\lambda_1, \lambda_2,\lambda_3, \lambda_4 \geq 0$ are tunable weighting hyperparameters (details can be found in Appendix~\ref{app:hyperparams}). $\mathcal{L}_{\text{text}}$ measures language generation quality via negative log-likelihood over a language dataset $\mathcal{D}_{\text{text}}$. $\mathcal{L}_{\text{act}}$ measures action prediction quality over a robot trajectory dataset $\mathcal{D}_{\text{act}}$; its form depends on the VLA action head — for an autoregressive architecture it is the negative log-likelihood over action tokens, while for flow matching it is the conditional flow matching loss~\cite{lipmanflow}. The KL divergence terms act as regularisers, penalising deviation from the pure VLM and VLA distributions respectively. Since $\theta_{\text{merged}}^{(l)}$ is a differentiable function of $\boldsymbol{\alpha}$ via Eq.~\eqref{eq:merge}, $\mathcal{L}$ is minimised directly with gradient descent using Adam \cite{kingma2017adammethodstochasticoptimization} (details can be found in Appendix~\ref{app:kl_terms}).

\section{Experiments}
 
\subsection{Experimental Setup}

\paragraph{Models and datasets.}
We apply our merging method to PaliGemma-3B~\cite{beyer2024paligemmaversatile3bvlm} as the VLM, paired with two VLA variants sharing the same architecture: $\pi_0$-FAST~\cite{pertsch2025fastefficientactiontokenization}, an autoregressive VLA that tokenises actions, and $\pi_{0.5}$~\cite{intelligence2025pi05visionlanguageactionmodelopenworld}, a flow matching VLA that predicts continuous action chunks via iterative denoising. Both VLA variants are obtained by fine-tuning PaliGemma on robot demonstration data. All models are kept frozen — only the mixing coefficients $\boldsymbol{\alpha}$ are optimised via Adam~\cite{kingma2017adammethodstochasticoptimization} on two complementary datasets built from the LIBERO dataset available on LeRobot~\cite{liu2023libero,cadene2026lerobotopensourcelibraryendtoend}. For the action signal, we relabel existing expert demonstrations from the LIBERO dataset with the pure VLA's own predicted action chunks, producing $\mathcal{D}_{\text{act}}$. For the language signal, we prompt PaliGemma to caption each observed frame, producing a paired dataset $\mathcal{D}_{\text{text}} = \{(\mathbf{x}_i, c_i)\}$ of observations and scene descriptions.
 \begin{table*}[ht]
\centering
\caption{Vision-language, embodied reasoning, and robot manipulation performance of all methods. The pure VLM cannot act, and both VLAs lose language and reasoning ability despite strong action performance. Existing merging baselines largely collapse on manipulation, while our merged models recover strong performance on all three axes simultaneously, without retraining either source model.}
\label{tab:vlm}
\resizebox{1.9\columnwidth}{!}{
\begin{tabular}{cccccccccccc}
\hline
\multicolumn{2}{c}{} & \multicolumn{4}{c}{\begin{tabular}[c]{@{}c@{}}VLM \\ Capabilities\end{tabular}} & \multicolumn{3}{c}{\begin{tabular}[c]{@{}c@{}}Embodied \\ Reasoning\end{tabular}} & \multicolumn{3}{c}{\begin{tabular}[c]{@{}c@{}}Robot \\ Manipulation\end{tabular}} \\ \hline
\multicolumn{1}{c|}{\multirow{2}{*}{\textbf{Model}}} & \multicolumn{1}{l|}{\multirow{2}{*}{Type}} & \multicolumn{3}{c}{Captioning} & \multicolumn{1}{c|}{VQAv2} & \multicolumn{3}{c|}{EQA} & \multicolumn{3}{c}{Libero} \\ \cline{3-12} 
\multicolumn{1}{c|}{} & \multicolumn{1}{l|}{} & RougeL & Meteor & BERTScore & \multicolumn{1}{c|}{Acc} & \begin{tabular}[c]{@{}c@{}}Obj\\ State\end{tabular} & Spatial & \multicolumn{1}{c|}{\begin{tabular}[c]{@{}c@{}}Attri\\  Recog\end{tabular}} & Spatial & Goal & Object \\ \hline
\multicolumn{1}{c|}{PaliGemma} & \multicolumn{1}{c|}{VLM} & \cellcolor[HTML]{40CC6C}33.6 & \cellcolor[HTML]{40CC6C}23.6 & \cellcolor[HTML]{40CC6C}88.9 & \multicolumn{1}{c|}{\cellcolor[HTML]{40CC6C}53.8} & \cellcolor[HTML]{40CC6C}44.4 & \cellcolor[HTML]{40CC6C}21.7 & \multicolumn{1}{c|}{\cellcolor[HTML]{40CC6C}32.4} & \cellcolor[HTML]{E74745}0.0 & \cellcolor[HTML]{E74745}0.0 & \cellcolor[HTML]{E74745}0.0 \\ \hline
\multicolumn{1}{c|}{$\pi_{0Fast}$} & \multicolumn{1}{c|}{\multirow{2}{*}{VLA}} & \cellcolor[HTML]{E74745}1.2 & \cellcolor[HTML]{E74745}0.5 & \cellcolor[HTML]{E74745}68.2 & \multicolumn{1}{c|}{\cellcolor[HTML]{E74745}9.9} & \cellcolor[HTML]{EF9942}15.5 & \cellcolor[HTML]{EC7E43}5.7 & \multicolumn{1}{c|}{\cellcolor[HTML]{EA6244}3.8} & \cellcolor[HTML]{88CE5A}70.0 & \cellcolor[HTML]{64CD63}82.0 & \cellcolor[HTML]{64CD63}88.0 \\
\multicolumn{1}{c|}{$\pi_{0.5}$} & \multicolumn{1}{c|}{} & \cellcolor[HTML]{EF9942}9.4 & \cellcolor[HTML]{E74745}1.5 & \cellcolor[HTML]{F1B540}76.5 & \multicolumn{1}{c|}{\cellcolor[HTML]{E74745}8.2} & \cellcolor[HTML]{E74745}8.8 & \cellcolor[HTML]{E74745}0 & \multicolumn{1}{c|}{\cellcolor[HTML]{E74745}0} & \cellcolor[HTML]{40CC6C}88.0 & \cellcolor[HTML]{40CC6C}90.0 & \cellcolor[HTML]{40CC6C}96.0 \\ \hline
\multicolumn{1}{c|}{Model Soup$_{\alpha=0.3}$} & \multicolumn{1}{c|}{\multirow{7}{*}{Merged}} & \cellcolor[HTML]{F4D03F}14.6 & \cellcolor[HTML]{F4D03F}11.4 & \cellcolor[HTML]{ACCF51}82.7 & \multicolumn{1}{c|}{\cellcolor[HTML]{D0CF48}32.4} & \cellcolor[HTML]{D0CF48}22.2 & \cellcolor[HTML]{88CE5A}15.1 & \multicolumn{1}{c|}{\cellcolor[HTML]{40CC6C}22.7} & \cellcolor[HTML]{E74745}0.0 & \cellcolor[HTML]{E74745}0.0 & \cellcolor[HTML]{E74745}0.0 \\
\multicolumn{1}{l|}{Model Soup$_{\alpha=0.5}$} & \multicolumn{1}{c|}{} & \cellcolor[HTML]{EC7E43}8.3 & \cellcolor[HTML]{EC7E43}5.3 & \cellcolor[HTML]{F1B540}76.4 & \multicolumn{1}{c|}{\cellcolor[HTML]{EF9942}21.7} & \cellcolor[HTML]{F4D03F}19.9 & \cellcolor[HTML]{EF9942}8.1 & \multicolumn{1}{c|}{\cellcolor[HTML]{F1B540}11.6} & \cellcolor[HTML]{E74745}0.0 & \cellcolor[HTML]{E74745}0.0 & \cellcolor[HTML]{E74745}0.0 \\
\multicolumn{1}{l|}{Model Soup$_{\alpha=0.7}$} & \multicolumn{1}{c|}{} & \cellcolor[HTML]{EA6244}5.0 & \cellcolor[HTML]{EC7E43}5.1 & \cellcolor[HTML]{EC7E43}73.9 & \multicolumn{1}{c|}{\cellcolor[HTML]{EC7E43}15.3} & \cellcolor[HTML]{F4D03F}19.9 & \cellcolor[HTML]{EA6244}4.4 & \multicolumn{1}{c|}{\cellcolor[HTML]{EA6244}4.7} & \cellcolor[HTML]{E74745}0.0 & \cellcolor[HTML]{E74745}0.0 & \cellcolor[HTML]{E74745}0.0 \\
\multicolumn{1}{l|}{Model Soup$_{\alpha=0.9}$} & \multicolumn{1}{c|}{} & \cellcolor[HTML]{E74745}2.1 & \cellcolor[HTML]{E74745}1.1 & \cellcolor[HTML]{E74745}70.1 & \multicolumn{1}{c|}{\cellcolor[HTML]{E74745}9.0} & \cellcolor[HTML]{EF9942}15.5 & \cellcolor[HTML]{EC7E43}5.8 & \multicolumn{1}{c|}{\cellcolor[HTML]{E74745}2.6} & \cellcolor[HTML]{EA6244}5.0 & \cellcolor[HTML]{EA6244}13.0 & \cellcolor[HTML]{EC7E43}17.0 \\
\multicolumn{1}{c|}{DARE} & \multicolumn{1}{c|}{} & \cellcolor[HTML]{EF9942}$10.4\pm0.9$ & \cellcolor[HTML]{EF9942}$7.6\pm0.4$ & \cellcolor[HTML]{F1B540}$78.3\pm0.3$ & \multicolumn{1}{c|}{\cellcolor[HTML]{EF9942}$22.5\pm1.7$} & \cellcolor[HTML]{E74745}$9.2\pm1.8$ & \cellcolor[HTML]{E74745}$3.6\pm0.9$ & \multicolumn{1}{c|}{\cellcolor[HTML]{EA6244}$4.1\pm0.8$} & \cellcolor[HTML]{E74745}$0.0\pm0.0$ & \cellcolor[HTML]{E74745}$0.0\pm0.0$ & \cellcolor[HTML]{E74745}$0.0\pm0.0$ \\
\multicolumn{1}{c|}{SLERP} & \multicolumn{1}{c|}{} & \cellcolor[HTML]{EF9942}9.9 & \cellcolor[HTML]{EF9942}7.4 & \cellcolor[HTML]{F1B540}77.7 & \multicolumn{1}{c|}{\cellcolor[HTML]{EC7E43}18.5} & \cellcolor[HTML]{F1B540}17.7 & \cellcolor[HTML]{F4D03F}10.3 & \multicolumn{1}{c|}{\cellcolor[HTML]{F1B540}10.5} & \cellcolor[HTML]{E74745}0.0 & \cellcolor[HTML]{E74745}0.0 & \cellcolor[HTML]{E74745}0.0 \\
\multicolumn{1}{c|}{AdaMerging$_{\pi0.5}$} & \multicolumn{1}{c|}{} & \cellcolor[HTML]{64CD63}$28.1\pm5.3$ & \cellcolor[HTML]{88CE5A}$23.1\pm5.8$ & \cellcolor[HTML]{64CD63}$87.5\pm1.1$ & \multicolumn{1}{c|}{\cellcolor[HTML]{88CE5A}$38.2\pm4.9$} & \cellcolor[HTML]{64CD63}$43.4\pm13.7$ & \cellcolor[HTML]{F4D03F}$14.0\pm3.8$ & \multicolumn{1}{c|}{\cellcolor[HTML]{88CE5A}$20.3\pm4.5$} & \cellcolor[HTML]{EA6244}$6.2\pm3.2$ & \cellcolor[HTML]{E74745}$2.9\pm1.1$ & \cellcolor[HTML]{E74745}$3.7\pm2.9$  \\
\hline
\multicolumn{1}{c|}{\textbf{Ours$_{\pi_{0Fast}}$}} & \multicolumn{1}{c|}{\textbf{Ours}} & \cellcolor[HTML]{D0CF48}$18.6\pm2.1$ & \cellcolor[HTML]{D0CF48}$14.1\pm1.1$ & \cellcolor[HTML]{ACCF51}$82.1\pm0.3$ & \multicolumn{1}{c|}{\cellcolor[HTML]{F4D03F}$28.9\pm1.9$} & \cellcolor[HTML]{64CD63}$29.7\pm4.8$ & \cellcolor[HTML]{88CE5A}$14.1\pm5.1$ & \multicolumn{1}{c|}{\cellcolor[HTML]{ACCF51}$17.2\pm6.9$} & \cellcolor[HTML]{ACCF51}$61.2\pm2.8$ & \cellcolor[HTML]{88CE5A}$73.1\pm3.5$ & \cellcolor[HTML]{64CD63}$79.8\pm3.1$ \\
\multicolumn{1}{c|}{\textbf{Ours$_{\pi_{0.5}}$}} & \multicolumn{1}{c|}{} & \cellcolor[HTML]{40CC6C}$29.5\pm1.8$ & \cellcolor[HTML]{40CC6C}$22.7\pm1.7$ & \cellcolor[HTML]{40CC6C}$87.8\pm0.5$ & \multicolumn{1}{c|}{\cellcolor[HTML]{40CC6C}$48.7\pm2.1$ } & \cellcolor[HTML]{40CC6C}$32.3\pm5.6$ & \cellcolor[HTML]{40CC6C}$17.3\pm6.4$ & \multicolumn{1}{c|}{\cellcolor[HTML]{88CE5A}$19.5\pm6.7$} & \cellcolor[HTML]{40CC6C}$89.3 \pm 3.1$ & \cellcolor[HTML]{40CC6C}$88.1 \pm 4.2$ & \cellcolor[HTML]{40CC6C}$93.3\pm2.6$ \\ \hline
\end{tabular}}

\vspace{0.2pt}
\newcommand{\lb}[2]{\colorbox[HTML]{#1}{\makebox[35pt][c]{\rule[-3pt]{0pt}{13pt}\footnotesize\textbf{#2}}}}
\begin{center}
{\footnotesize}
{Score: \setlength{\fboxsep}{0pt}%
\lb{E74745}{Low}\lb{EA6244}{}\lb{EC7E43}{}\lb{EF9942}{}\lb{F1B540}{}\lb{F4D03F}{Medium}\lb{D0CF48}{}\lb{ACCF51}{}\lb{88CE5A}{}\lb{64CD63}{}\lb{40CC6C}{High}}\\[3pt]

\end{center}

\end{table*}
\paragraph{Evaluation protocol.}
We evaluate the merged model along three axes: vision-language capabilities, embodied reasoning, and action capabilities.

For \textbf{vision-language capabilities}, we assess captioning performance on the COCO dataset~\cite{lin2015microsoftcococommonobjects} using Rouge-L~\cite{lin-2004-rouge}, METEOR~\citep{banerjee-lavie-2005-meteor} and BERTScore~\citep{zhang2020bertscoreevaluatingtextgeneration} metrics, and visual question answering performance on VQAv2~\cite{balanced_vqa_v2}.

For \textbf{embodied reasoning}, we evaluate accuracy on the EQA benchmark across three question categories — Object State, Spatial, and Attribute Recognition — which probe whether the model can answer language questions grounded in a robotic setting, bridging the gap between the general-domain vision-language evaluations above and the action benchmarks below.

For \textbf{action capabilities}, we evaluate task success rate on the LIBERO task suites — LIBERO-Spatial, LIBERO-Goal and LIBERO-Object~\cite{liu2023libero} — and report the average success rate across tasks for each suite (details can be found in Appendix~\ref{app:envs}; additional results on the RoboCasa kitchen tasks are also reported in Appendix~\ref{app:robocasa}).
\paragraph{Baselines.}
We compare our merged model against a set of baselines. The \textbf{VLM} (PaliGemma~\cite{beyer2024paligemmaversatile3bvlm}) and \textbf{VLAs} ($\pi_0$-FAST and $\pi_{0.5}$) serve as upper bounds on language and action performance respectively. $\pi_{0.5}$~\cite{intelligence2025pi05visionlanguageactionmodelopenworld} is also a co-trained model that jointly optimises language and action objectives during training. We further include three established merging strategies that all use a fixed, non-learned merging coefficient: \textbf{Model Soup}~\cite{wortsman2022modelsoupsaveragingweights}, which linearly interpolates between VLM and VLA weights, and which we evaluate at four interpolation coefficients $\alpha \in \{0.3, 0.5, 0.7, 0.9\}$ to characterise the full trade-off curve it induces between language and action performance; \textbf{DARE}~\cite{yu2024languagemodelssupermario}, which randomly drops a fraction of the task vector parameters before rescaling and adding the remainder to the VLM; and \textbf{SLERP}~\cite{jang2024sphericallinearinterpolationtextanchoring}, which interpolates between the two weight vectors along the spherical geodesic rather than linearly. For the merged baselines, $\pi_{0FAST}$ is used as the VLA backbone.
We additionally compare against \textbf{AdaMerging}~\cite{yang2024adamergingadaptivemodelmerging}, which learns a single per-layer coefficient rather than per-component ones; we optimise it with the same objective (Eq.~\ref{eq:objective}) as our method; we use $\pi_{0.5}$ as its VLA backbone for direct comparison with \textbf{Ours}$_{\pi_{0.5}}$.
All stochastic methods — DARE, AdaMerging, and our merged models — are averaged over 10 seeds, with standard deviations reported in Table~\ref{tab:vlm}.
\subsection{Main Results}
\begin{figure*}[htbp]
    \centering
    \includegraphics[width=0.95\linewidth]{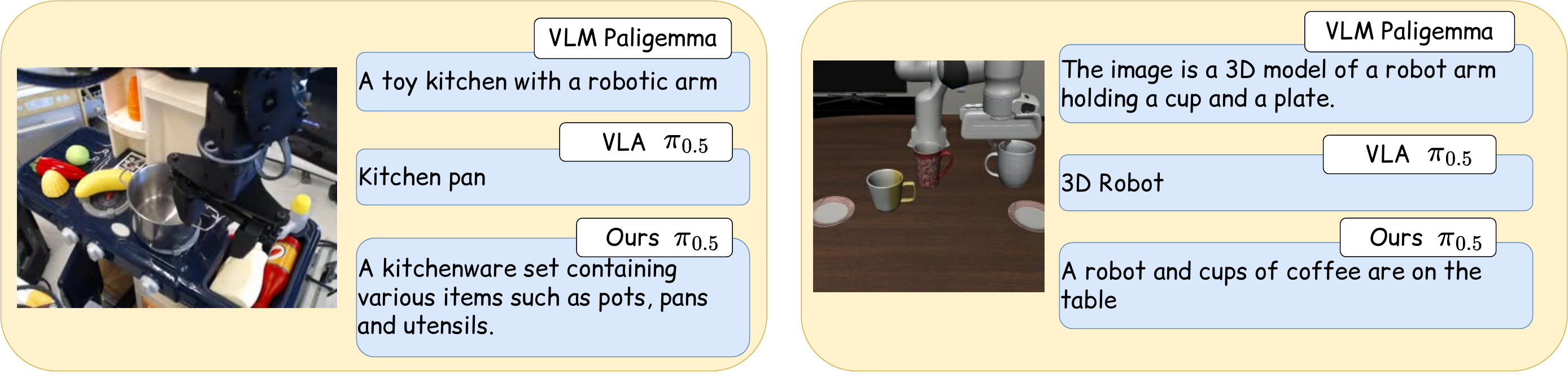}
    \caption{\textbf{Qualitative captioning examples.} Given the same visual observation, the VLM produces coherent scene descriptions, the VLA ($\pi_{0.5}$) generates fragmented or incomplete text due to catastrophic forgetting, and \textbf{ Ours}$_{\pi_{0.5}}$ merged model recovers meaningful captions while retaining action capabilities.}
    \label{fig:qualitative}
    \vspace{-0.5cm}
\end{figure*}
\paragraph{Vision-language capabilities.}
\label{sec:vlm}
Table~\ref{tab:vlm} reports captioning and VQA performance. The pure VLM sets the upper bound (33.68 ROUGE-L, 88.98 BERTScore, 53.87 VQAv2), while both pure VLAs degrade severely ($\pi_0$-FAST: 1.25 ROUGE-L, 9.93 VQAv2; $\pi_{0.5}$: 9.40 ROUGE-L, 8.22 VQAv2 despite co-training). Model Soup, DARE and SLERP offer only modest, comparable recovery (8.32--10.11 ROUGE-L), confirming that a single coefficient applied uniformly across the task vector cannot restore language grounding. AdaMerging$_{\pi_{0.5}}$, which shares our training objective but learns only a per-layer coefficient, recovers substantially more (28.1 ROUGE-L, 87.5 BERTScore, 38.2 VQAv2) — but still trails\textbf{ Ours}$_{\pi_{0.5}}$ (29.5/87.8/48.7), and, as shown below, this comes at the cost of near-total collapse on action prediction. Our merged models close most of the remaining gap to the VLM: Ours$_{\pi_{0Fast}}$ reaches 18.6 ROUGE-L / 28.9 VQAv2, and Ours$_{\pi_{0.5}}$ reaches 29.5 ROUGE-L / 87.8 BERTScore / 48.7 VQAv2 — nearly matching the VLM. Recovery is stronger for $\pi_{0.5}$ than $\pi_0$-FAST, likely because the latter entangles language and action within a shared autoregressive head. Figure~\ref{fig:qualitative} illustrates this: $\pi_{0.5}$ retains scene-relevant keywords (e.g.\ ``Kitchen pan'', ``Robot'') but loses the language modelling coherence of the VLM needed to form descriptions, whereas our merged model generates grammatically coherent captions that correctly identify objects and their spatial relationships.
Additional examples can be found in Appendix~\ref{app:qualitative}

\paragraph{Embodied reasoning.}
The same pattern holds on EQA (Object State / Spatial / Attribute Recognition). The VLM leads (44.44/21.73/32.46), both VLAs degrade sharply, and fixed-coefficient baselines recover only partially and inconsistently (best: Model Soup$_{\alpha=0.3}$ at 22.22/15.19/22.75). AdaMerging$_{\pi_{0.5}}$ recovers markedly more (43.4/14.0/20.3), even edging out\textbf{ Ours}$_{\pi_{0.5}}$ on Object State, indicating that a learned, task-aware objective is the primary driver of embodied-reasoning recovery, with SVD granularity offering more modest further gains on this axis. Our merged models nonetheless outperform all baselines overall:\textbf{ Ours}$_{\pi_{0Fast}}$ reaches 29.7/14.1/17.2, and\textbf{ Ours}$_{\pi_{0.5}}$ reaches 32.3/17.3/19.5, recovering a substantial share of the VLM's embodied reasoning capabilities despite having no direct EQA supervision.
\begin{figure*}[t]
    \centering
    \includegraphics[width=0.92\linewidth]{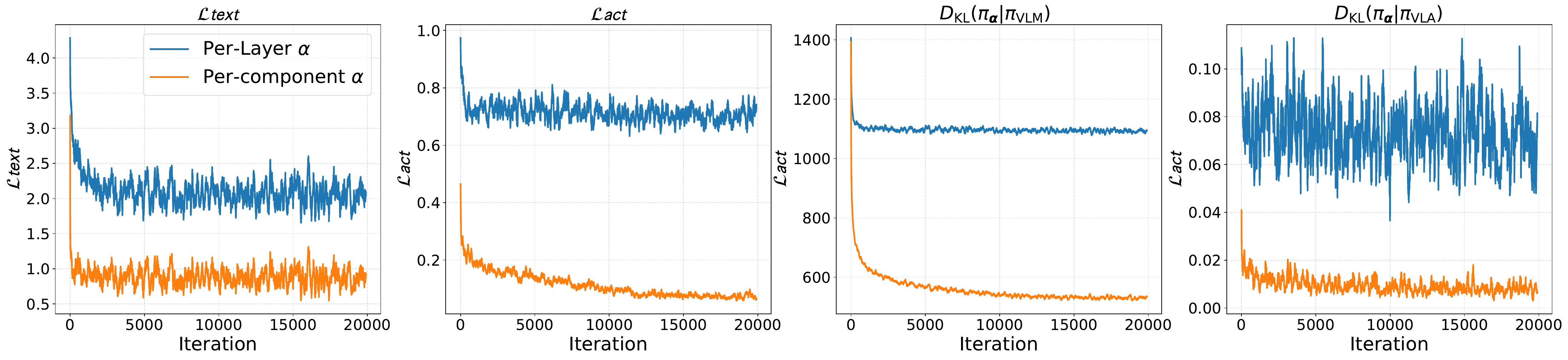}
    \caption{\textbf{Granularity of mixing coefficients.} We compare two levels of granularity for $\boldsymbol{\alpha}$ — per-layer and per-component (ours) — and report the evolution of $\mathcal{L}_{\text{text}}$, $\mathcal{L}_{\text{act}}$, $D_{\text{KL}}(\pi_{\boldsymbol{\alpha}} \| \pi_{\text{VLM}})$ and $D_{\text{KL}}(\pi_{\boldsymbol{\alpha}} \| \pi_{\text{VLA}})$ over training. Per-component $\boldsymbol{\alpha}$ converges to substantially lower values on every metric, confirming that fine-grained SVD-based control is essential to simultaneously preserve language and action capabilities.}
    \label{fig:training_curves}
    \vspace{-0.4cm}
\end{figure*}
\paragraph{Robot manipulation performance.}
Table~\ref{tab:vlm} also reports success rate on the LIBERO suites. The VLM achieves zero success everywhere, while both VLAs establish strong upper bounds ($\pi_{0.5}$: 88/90/96; $\pi_0$-FAST: 70/82/88). DARE, SLERP and Model Soup (at $\alpha \leq 0.7$) collapse entirely to zero success, and Model Soup$_{\alpha=0.9}$ only recovers marginally (5/13/17) by sacrificing nearly all language gains — confirming that a fixed, uniform coefficient cannot preserve the action-relevant structure of the VLA. AdaMerging$_{\pi_{0.5}}$ collapses almost as severely (6.2/2.9/3.7) despite sharing our training objective, showing that a single per-layer coefficient lacks the resolution to preserve action-relevant directions — consistent with the granularity ablation of Section~\ref{sec:ablation}, where per-layer $\boldsymbol{\alpha}$ cannot disentangle language- from action-relevant directions within $\Delta^{(l)}$. Our merged models retain strong performance on both variants:\textbf{ Ours}$_{\pi_{0Fast}}$ reaches 61.2/73.1/79.8, and\textbf{ Ours}$_{\pi_{0.5}}$ reaches 89.3/88.1/93.3, closely tracking its pure VLA counterpart while still achieving the language and reasoning gains above — a trade-off no baseline, including AdaMerging, offers.
\subsection{Ablation study}

\label{sec:ablation}
We conduct three ablations to validate the design choices of our method: the granularity of the mixing coefficients, the contribution of each component of the training objective, and the sensitivity of the method to the rank of the SVD decomposition.
\paragraph{Granularity of mixing coefficients.}

To analyse the impact of per-component SVD coefficients, we compare two levels of granularity for $\boldsymbol{\alpha}$. Under \textbf{per-layer} $\boldsymbol{\alpha}$, a single coefficient $\alpha^{(l)} \in [0,1]$ is learned for each layer $l$ and applied uniformly to every singular component of that layer's task vector $\Delta^{(l)}$ — this is equivalent to the layer-wise \textbf{AdaMerging }protocol~\cite{yang2024adamergingadaptivemodelmerging}, adapted to our setting. Under \textbf{per-component} $\boldsymbol{\alpha}$ (ours), a separate coefficient $\alpha_k^{(l)} \in [0,1]$ is learned for each individual singular component $k$ within each layer $l$. Figure~\ref{fig:training_curves} shows the evolution of $\mathcal{L}_{\text{text}}$, $\mathcal{L}_{\text{act}}$, $D_{\text{KL}}(\pi_{\boldsymbol{\alpha}} \| \pi_{\text{VLM}})$ and $D_{\text{KL}}(\pi_{\boldsymbol{\alpha}} \| \pi_{\text{VLA}})$ over optimisation steps for each condition. The results are unambiguous: per-component $\boldsymbol{\alpha}$ consistently and substantially outperforms the per-layer scalar across all four metrics. On $\mathcal{L}_{\text{text}}$, per-layer $\boldsymbol{\alpha}$ plateaus around 2.0, while per-component $\boldsymbol{\alpha}$ converges below 1.0. The gap is even more pronounced on action prediction: per-layer $\boldsymbol{\alpha}$ converges to $\mathcal{L}_{\text{act}} \approx 0.7$ and $D_{\text{KL}}(\pi_{\boldsymbol{\alpha}} \| \pi_{\text{VLA}})$ remains noisy around 0.06--0.08, whereas per-component $\boldsymbol{\alpha}$ drives both metrics down to roughly a tenth of these values. A single coefficient per layer cannot disentangle the directions within $\Delta^{(l)}$ that encode language degradation from those that encode action prediction, since both are entangled within the same weight matrix and a layer-wide $\alpha^{(l)}$ must trade them off uniformly. These results confirm that the fine-grained control afforded by SVD decomposition is essential — coarser parameterisations lack the expressivity to reconcile both objectives.
\paragraph{Contribution of each loss component.}
\begin{table}[t]
\centering
\caption{Contribution of each loss component. We ablate each term of the training objective and evaluate using ROUGE-L for language quality and Success Rate SR for action predictions. Best results in \textbf{bold}.}
\label{tab:ablation_loss}
\resizebox{0.75\columnwidth}{!}{
\begin{tabular}{lcc}
\toprule
\textbf{Objective} & ROUGE-L $\uparrow$ & SR $\uparrow$ \\
\midrule
w/o $\mathcal{L}_{\text{text}}$                                     & 21.9             & 95.6 \\
w/o $\mathcal{L}_{\text{act}}$                                      & 34.6          & 0     \\
w/o $D_{\text{KL}}(\pi_{\boldsymbol{\alpha}} \| \pi_{\text{VLM}})$ & 7.9          & 94.1 \\
w/o $D_{\text{KL}}(\pi_{\boldsymbol{\alpha}} \| \pi_{\text{VLA}})$ & 31.2          & 71.5 \\
\midrule
Full objective (ours)                                               & \textbf{29.5} & \textbf{93.3} \\
\bottomrule

\end{tabular}}
\vspace{-0.5cm}
\end{table}

 \begin{figure*}[t]
\centering
    \includegraphics[width=0.46\linewidth]{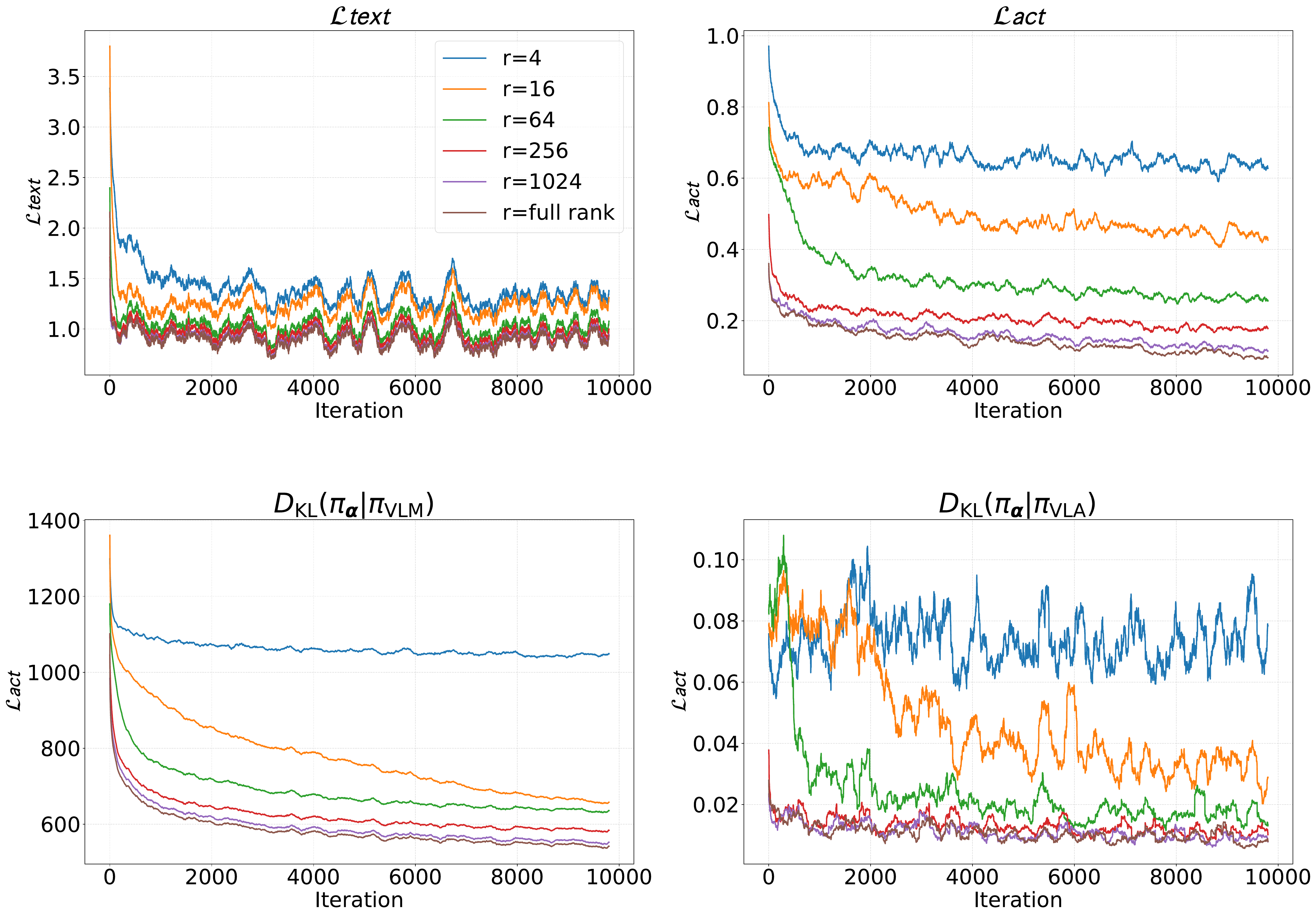} \includegraphics[width=0.46\linewidth]{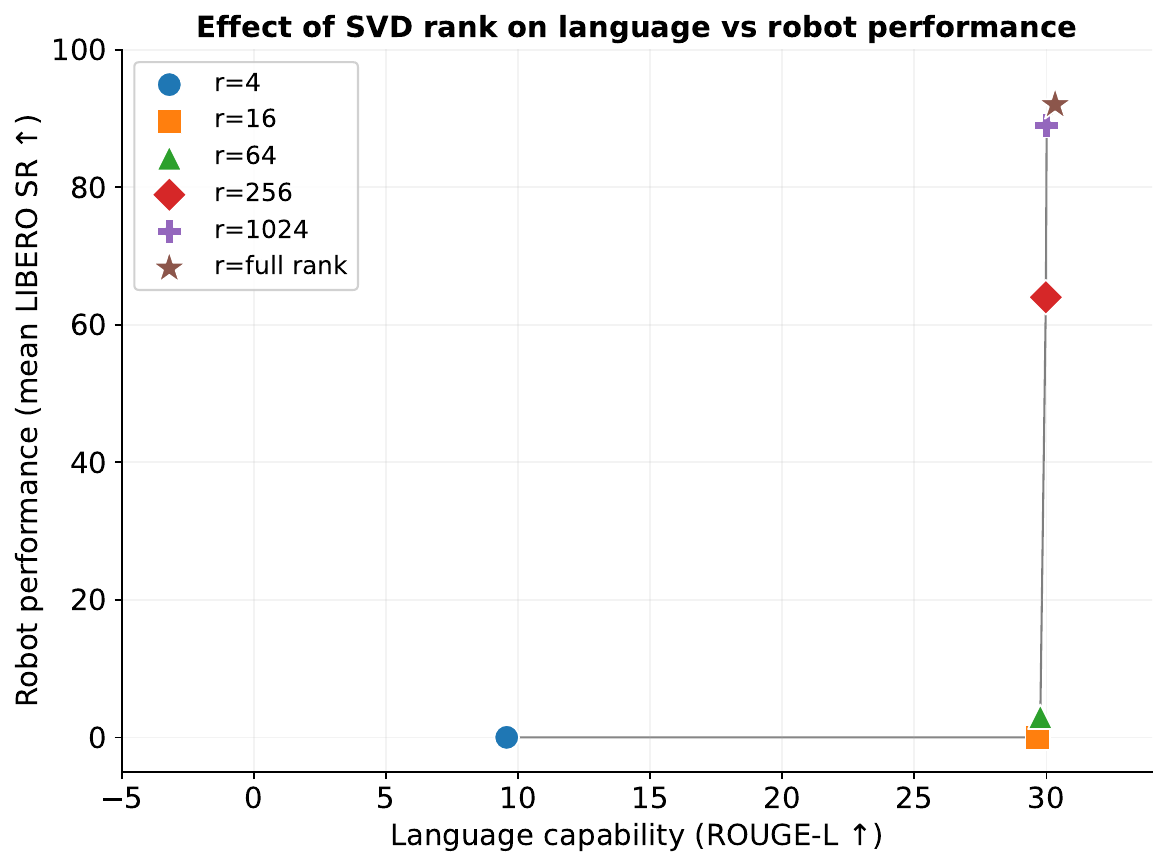}
    \caption{\textbf{Effect of SVD rank.} \textit{(Left)} Training losses for varying rank $r \in \{4, 16, 64, 256, 1024, \text{full}\}$: $\mathcal{L}_{\text{text}}$ is largely insensitive to rank, while $\mathcal{L}_{\text{act}}$ improves markedly with higher rank, indicating that the VLA adaptation is diffuse across many singular components. \textit{(Right)} Downstream evaluation confirms this asymmetry: ROUGE-L saturates from $r=16$, while LIBERO success rate requires $r \geq 1024$ to exceed 90\% — language recovery needs far fewer singular directions than action preservation.}
    \label{fig:rank_sweep}
    \vspace{-0.6cm}
\end{figure*} 
To understand the role of each term in the training objective $\mathcal{L}(\boldsymbol{\alpha})$, we train four variants of\textbf{ Ours}$_{\pi_{0.5}}$ by removing one component at a time: without $\mathcal{L}_{\text{text}}$, $\mathcal{L}_{\text{act}}$, $D_{\text{KL}}(\pi_{\boldsymbol{\alpha}} \| \pi_{\text{VLM}})$, and $D_{\text{KL}}(\pi_{\boldsymbol{\alpha}} \| \pi_{\text{VLA}})$. Results are reported in Table~\ref{tab:ablation_loss}. Removing $\mathcal{L}_{\text{act}}$ causes the merged model to collapse on action prediction entirely (0\% success rate) while language is the highest of all variants (34.6 ROUGE-L), confirming that without an explicit action signal $\boldsymbol{\alpha}$ drifts toward the VLM. Conversely, removing $D_{\text{KL}}(\pi_{\boldsymbol{\alpha}} \| \pi_{\text{VLM}})$ severely degrades language quality (7.9 ROUGE-L) while leaving action performance almost intact (94.1\%), showing that this KL term is the primary anchor preserving language fluency and that $\mathcal{L}_{\text{text}}$ alone is insufficient to fully constrain $\boldsymbol{\alpha}$ toward the VLM. Removing $\mathcal{L}_{\text{text}}$ has a comparatively mild effect (21.97 ROUGE-L, 95\% success rate), suggesting that $D_{\text{KL}}(\pi_{\boldsymbol{\alpha}} \| \pi_{\text{VLM}})$ already provides most of the language-preserving signal on its own. Removing $D_{\text{KL}}(\pi_{\boldsymbol{\alpha}} \| \pi_{\text{VLA}})$ yields the second-best language score (31.2 ROUGE-L) but a substantial drop in success rate (71.5\%), indicating that this term plays an important role in stabilising action prediction beyond what $\mathcal{L}_{\text{act}}$ alone provides. Overall, the full objective achieves the best joint trade-off (29.5 ROUGE-L, 93.3\% success rate), confirming that all four terms are complementary: the $\cal{L}$ terms provide the primary task signal while the $KL$ terms regularise and stabilise the optimisation, with the VLM-side KL term proving especially critical for language retention.


\paragraph{Effect of SVD rank.}

We study the sensitivity of the method to the rank $r$ used in the SVD decomposition by varying $r \in \{4, 16, 64, 256, 1024, \text{full}\}$. For a given $r$, each layer's task vector $\Delta^{(l)}$ is truncated to its top-$r$ singular components — those associated with the $r$ largest singular values — and only the corresponding $\alpha_k^{(l)}$ are optimised, with the remaining components implicitly set to the VLM. Figure~\ref{fig:rank_sweep} reveals a striking asymmetry between the two objectives: $\mathcal{L}_{\text{text}}$ is largely insensitive to rank, while $\mathcal{L}_{\text{act}}$ shows a clear separation, with higher rank leading to significantly better action preservation. The KL divergence terms follow the same pattern. This is confirmed by the downstream scatter plot: ROUGE-L is low at the extreme $r=4$ but already saturates around 30 from $r=16$ onward, while mean LIBERO success rate stays at 0\% for $r \leq 16$, jumps to 64\% at $r=256$, and exceeds 90\% only from $r=1024$ onward. Notably, $r=1024$ requires only 269{,}443 learnable parameters compared to 381{,}605 at full rank — both roughly four orders of magnitude smaller than the finetuning of the $\sim$3B backbone, confirming that near-optimal performance is achieved with an even smaller fraction of the already lightweight parameter budget. This indicates that language recovery only requires a small set of dominant directions of the task vector, whereas action preservation requires a much larger, near complete, set of singular directions — confirming that the VLA adaptation is diffuse in weight space while language-relevant directions are concentrated in a handful of dominant components. More broadly, this asymmetry explains why uniform merging fails to recover action performance while partially recovering language. More analysis on the $\alpha$ distribution can be found in Appendix~\ref{app:alpha_rank}
\section{Conclusion}
 
In this paper, we present a post-hoc method for merging a VLM and a VLA by decomposing the task vector via SVD and learning per-component mixing coefficients through gradient descent. We apply this method to two distinct VLA architectures, an autoregressive and a flow matching model, demonstrating that it generalises across different action decoding strategies. Both source models remain frozen throughout, and only the mixing coefficients are optimised, making the approach substantially cheaper than retraining or co-training a VLA from scratch and broadly applicable to already fine-tuned VLAs. Our approach recovers a model that retains the action prediction capabilities of the VLA while restoring the vision-language understanding of the original VLM, a balance that uniform interpolation fails to achieve. Through extensive ablations, we show that this fine-grained control over the merging process is essential, and our analysis further reveals that language and action capabilities are encoded at different scales in weight space, offering new insight into the nature of catastrophic forgetting in VLA fine-tuning. Future work could leverage the language and reasoning capabilities restored by our method to equip VLAs with high-level planning, enabling them to decompose and reason about long-horizon tasks rather than merely reacting to immediate instructions. Code and merged model checkpoints will be released upon acceptance.
\section*{Limitations}

Our method assumes the VLM and VLA share a common backbone, so it does not directly apply to mismatched architectures or tokenizers. It also still requires a training loop to optimise $\boldsymbol{\alpha}$, albeit a much cheaper one than full fine-tuning. Our action and language datasets are both derived from a single benchmark (LIBERO) and from the source models themselves, so performance on out-of-distribution tasks, embodiments, or real-world deployments remains untested. Finally, our experiments are limited to VLMs and VLAs at the 3--4B parameter scale, and whether our findings hold at larger scales remains untested.

\section*{Ethical Considerations}

This work aims to mitigate catastrophic forgetting in robot learning models, helping robotic agents retain language-based communication alongside low-level control rather than sacrificing one for the other. All experiments are conducted in simulation using publicly available models and datasets; no human subjects or physical robot deployments are involved. As with any method that enhances action capabilities, thorough safety validation is recommended before real-world deployment.

\section*{Acknowledgments}
Experiments presented in this paper were carried out using the HPC resources of IDRIS under the allocations 2026-[A0201016980], 2026-[AD011015093R2] and 2026-[A0181015978] made by GENCI. This work was supported by the European Commission's Horizon Europe Framework Programme under grant No 101070381 (PILLAR-robots), by RODEO Project (ANR-24-CE23-5886), by PEPR Sharp (ANR-23-PEIA-0008, ANR,
FRANCE 2030) and by Cluster PostGenAI@Paris (ANR-23-IACL-0007, FRANCE 2030). 
\bibliography{custom}

\newpage 
\appendix

\onecolumn
\section{Implementation Details}
\subsection{Hyperparameters}
\label{app:hyperparams}
Table~\ref{tab:hyperparams} reports the full set of hyperparameters used to train the mixing coefficients $\boldsymbol{\alpha}$ for our main results, separately for\textbf{ Ours}$_{\pi_{0Fast}}$ and\textbf{ Ours}$_{\pi_{0.5}}$.
\begin{table}[ht]
\centering
\caption{Hyperparameters used to optimise $\boldsymbol{\alpha}$ for each VLA variant. $\lambda_1,\lambda_2$ weight $\mathcal{L}_{\text{text}}$ and $\mathcal{L}_{\text{act}}$ respectively; $\lambda_3,\lambda_4$ weight the KL regularisation terms toward the VLM and VLA respectively.}
\label{tab:hyperparams}
\begin{tabular}{lcc}
\toprule
\textbf{Hyperparameter} & \textbf{Ours$_{\pi_{0Fast}}$} & \textbf{Ours$_{\pi_{0.5}}$} \\
\midrule
Optimiser                          & Adam            & Adam \\
Initial learning rate              & $0.01$          & $0.01$ \\
Learning rate schedule             & cosine decay    & cosine decay \\
Total training steps               & 20{,}000        & 20{,}000 \\
$\lambda_1$ ($\mathcal{L}_{\text{text}}$) & 1        & 1 \\
$\lambda_2$ ($\mathcal{L}_{\text{act}}$)  & 1        & 5 \\
$\lambda_3$ (KL$_{\text{VLM}}$)    & 0.01            & 0.01 \\
$\lambda_4$ (KL$_{\text{VLA}}$)    & 0.8              & 10 \\
$\boldsymbol{\alpha}$ initialisation & Uniform$(0,1)$ & Uniform$(0,1)$ \\
SVD power iterations               & 4               & 4 \\
Batch size                         & 2               & 2 \\
Number of GPUs                     & 1               & 1 \\
\bottomrule
\end{tabular}
\end{table}
\subsection{KL Regularisation Terms}
\label{app:kl_terms}
 
\paragraph{Language side.} For the VLM side, $\pi_{\text{VLM}}$ and $\pi_{\boldsymbol{\alpha}}$ denote the token-level output distributions of the pure VLM and the merged model, respectively, computed over the language vocabulary at each generated position given the same visual observation and prompt. For $\pi_0$-FAST, whose vocabulary jointly contains both language and action tokens, this distribution is restricted to the language sub-vocabulary, masking out action-token logits before computing the divergence. $D_{\text{KL}}(\pi_{\boldsymbol{\alpha}} \| \pi_{\text{VLM}})$ is the standard categorical KL divergence between these two (vocabulary-restricted) token distributions, averaged over sequence positions, and is well-defined for both VLA variants since the language head is shared and discrete in both $\pi_0$-FAST and $\pi_{0.5}$.
 
\paragraph{Action side — autoregressive VLA.} For $\pi_0$-FAST, actions are represented as discrete tokens drawn from a dedicated action sub-vocabulary, and $\pi_{\text{VLA}}$, $\pi_{\boldsymbol{\alpha}}$ denote the corresponding token-level distributions restricted to this action sub-vocabulary, with language-token logits masked out. $D_{\text{KL}}(\pi_{\boldsymbol{\alpha}} \| \pi_{\text{VLA}})$ is therefore also a standard categorical KL divergence, computed analogously to the language term above but over the disjoint action sub-vocabulary, ensuring that each KL term reflects only the corresponding modality.
 
\paragraph{Action side — flow matching VLA.} For $\pi_{0.5}$, actions are continuous and generated via iterative denoising, so no closed-form categorical distribution exists over the action space. Following prior work on distribution matching for diffusion- and flow-based visuomotor policies~\cite{jia2024scoredistributionmatchingpolicy,ping2026flowdppodivergenceproximalpolicy}, we approximate $\pi_{\text{VLA}}$ and $\pi_{\boldsymbol{\alpha}}$ as Gaussian distributions over the predicted velocity field. Concretely, for a given observation we sample $N=16$ independent noise draws $x_0^{(1)}, \dots, x_0^{(N)} \sim \mathcal{N}(0, I)$, form the corresponding interpolated actions $x_t^{(i)} = t\,x_1 + (1-t)\,x_0^{(i)}$, and compute the predicted velocities $v_{\theta_{\text{VLA}}}^{(i)}$ and $v_{\theta_{\text{merged}}}^{(i)}$ under the pure VLA and the merged model respectively. We then estimate the empirical mean and variance of each set of velocity predictions,
\begin{equation}
    \mu_{\theta} = \frac{1}{N} \sum_{i=1}^N v_{\theta}^{(i)}, \qquad
    \sigma_{\theta}^2 = \frac{1}{N} \sum_{i=1}^N \left(v_{\theta}^{(i)} - \mu_{\theta}\right)^2, \qquad \theta \in \{\theta_{\text{VLA}}, \theta_{\text{merged}}\},
\end{equation}
and approximate $\pi_{\text{VLA}} \approx \mathcal{N}(\mu_{\theta_{\text{VLA}}}, \sigma_{\theta_{\text{VLA}}}^2)$, $\pi_{\boldsymbol{\alpha}} \approx \mathcal{N}(\mu_{\theta_{\text{merged}}}, \sigma_{\theta_{\text{merged}}}^2)$, for which the KL divergence admits the standard closed form:
\begin{equation}
    D_{\text{KL}}(\pi_{\boldsymbol{\alpha}} \| \pi_{\text{VLA}}) = \frac{1}{2}\left(
        \frac{\sigma_{\theta_{\text{merged}}}^2}{\sigma_{\theta_{\text{VLA}}}^2}
        + \frac{(\mu_{\theta_{\text{VLA}}} - \mu_{\theta_{\text{merged}}})^2}{\sigma_{\theta_{\text{VLA}}}^2}
        - 1
        + \log\frac{\sigma_{\theta_{\text{VLA}}}^2}{\sigma_{\theta_{\text{merged}}}^2}
    \right).
\end{equation}
This Gaussian approximation is computed independently per action dimension and averaged. We note that this is an approximation of the true (intractable) distribution induced by the iterative denoising process, justified by the fact that the velocity field at a fixed timestep is approximately unimodal in practice.
\subsection{Dataset Collection}
\label{app:dataset_collection}
 
\paragraph{Action dataset $\mathcal{D}_{\text{act}}$.} For each VLA variant, we construct $\mathcal{D}_{\text{act}}$ by relabeling existing expert demonstrations from the LIBERO dataset available on LeRobot~\cite{liu2023libero,cadene2026lerobotopensourcelibraryendtoend}. Concretely, we take the visual observations, proprioceptive states, and task instructions from the expert trajectories and query the pure VLA to obtain its predicted action chunks for each observation, storing the resulting pairs.  This collection is performed once per VLA variant and reused throughout training and ablation, and yields a total of 5{,}000 examples.
 
\paragraph{Language dataset $\mathcal{D}_{\text{text}}$.} For each visual observation collected above, we additionally query the pure VLM with a captioning prompt to obtain a corresponding scene description, producing paired examples $(\mathbf{x}_i, c_i) \in \mathcal{D}_{\text{text}}$, also totalling 5{,}000 examples. Both $\mathcal{D}_{\text{act}}$ and $\mathcal{D}_{\text{text}}$ are split into disjoint train and test partitions, and all reported metrics are computed on test examples.
 
\subsection{Training Cost}
\label{app:training_cost}
Training the mixing coefficients $\boldsymbol{\alpha}$ for our main results takes approximately 3 hours on a single NVIDIA H100 GPU for 20{,}000 steps. Both source models ($\theta_{\text{VLM}}$ and $\theta_{\text{VLA}}$) remain frozen throughout, and only $\boldsymbol{\alpha}$ is updated. At our chosen practical rank, this amounts to a total of 381{,}587 learnable parameters for\textbf{ Ours}$_{\pi_{0Fast}}$ and 381{,}605 for\textbf{ Ours}$_{\pi_{0.5}}$ across all layers weight matrices — in both cases roughly four orders of magnitude smaller than the $\sim$3B frozen parameters of the VLA backbone alone. This makes our approach substantially cheaper than retraining or co-training a VLA from scratch.

\section{Environment and Task Definitions}
\label{app:envs}
\paragraph{LIBERO.} LIBERO~\cite{liu2023libero} is a simulation benchmark for lifelong robot manipulation built on the Robosuite simulator, comprising several task suites each containing 10 tasks with 50 expert demonstrations per task. We evaluate on three of its suites: \textbf{LIBERO-Spatial}, where the robot must place a bowl on a plate among visually identical bowls that differ only in their spatial location relative to other objects, testing spatial reasoning under a fixed set of objects; \textbf{LIBERO-Object}, where the robot must pick and place a different, unique target object into a basket across tasks while the layout remains fixed, testing object-centric generalisation; and \textbf{LIBERO-Goal}, where both the objects and their spatial layout are held fixed while the task goal varies, testing the transfer of procedural, goal-directed behaviour. Each suite is designed to disentangle a specific axis of generalisation (spatial, object, or goal knowledge), and observations consist of third-person and wrist-camera RGB images.
\begin{table}[ht]
\centering
\caption{Task success rate (\%) on three RoboCasa kitchen tasks. All fixed-coefficient merging baselines collapse to zero, while our merged models retain meaningful performance on both VLA variants.}
\label{tab:robocasa}
\begin{tabular}{c|c|ccc}
\toprule
\textbf{Model} & \textbf{Type} & \begin{tabular}[c]{@{}c@{}}Open\\ Fridge\end{tabular} & \begin{tabular}[c]{@{}c@{}}Setup\\ Cup\end{tabular} & \begin{tabular}[c]{@{}c@{}}Navigate\\ Kitchen\end{tabular} \\
\midrule
PaliGemma & VLM & 0 & 0 & 0 \\
\midrule
$\pi_{0Fast}$ & \multirow{2}{*}{VLA} & 78 & 68 & 89 \\
$\pi_{0.5}$ &  & 71 & 58 & 25 \\
\midrule
Model Soup$_{\alpha=0.3}$ & \multirow{8}{*}{Merged} & 0 & 0 & 0 \\
Model Soup$_{\alpha=0.5}$ &  & 0 & 0 & 0 \\
Model Soup$_{\alpha=0.7}$ &  & 0 & 0 & 0 \\
Model Soup$_{\alpha=0.9}$ &  & 0 & 0 & 0 \\
DARE &  & 0 & 0 & 0 \\
SLERP &  & 0 & 0 & 0 \\
\textbf{Ours$_{\pi_{0Fast}}$} &  & 56 & 61 & 78 \\
\textbf{Ours$_{\pi_{0.5}}$} &  & 65 & 49 & 21 \\
\bottomrule
\end{tabular}
\end{table}
 \begin{figure*}[ht]
    \centering
    $\mathbf{Ours}_{\pi_{0FAST}}$\\\includegraphics[width=\linewidth]{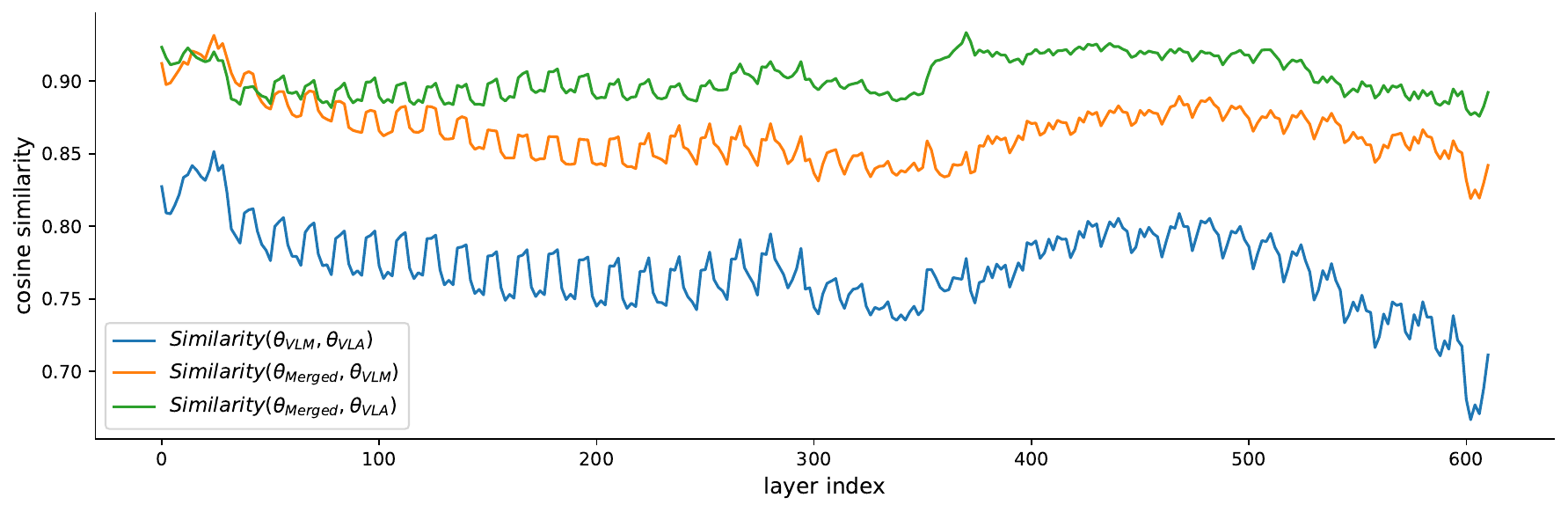}\\$\mathbf{Ours}_{\pi_{0.5}}$\\\includegraphics[width=\linewidth]{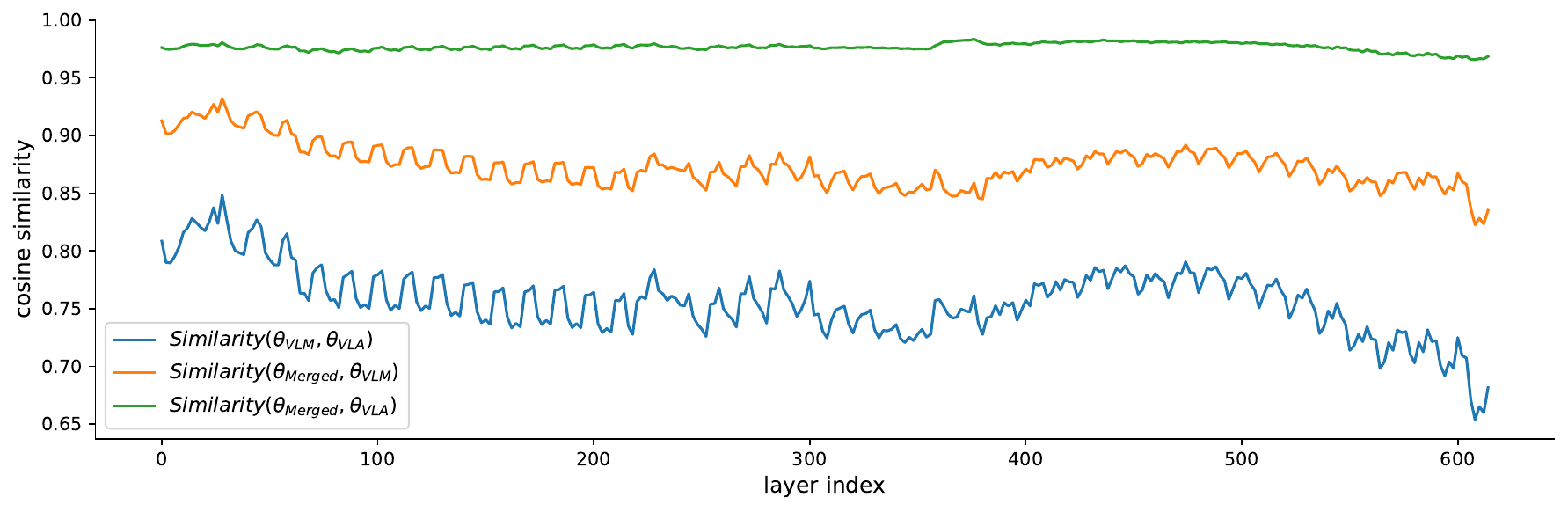}
    \caption{\textbf{Per-layer similarity to source models.} \textit{(Top)}\textbf{ Ours}$_{\pi_{0Fast}}$. \textit{(Bottom)}\textbf{ Ours}$_{\pi_{0.5}}$. For each of the 602 weight matrices, we compute the cosine similarity between the VLM and VLA weights (blue), between the merged model and the VLM (orange), and between the merged model and the VLA (green). In both cases, the merged model is consistently closer to the VLA than to the VLM across all layers, while remaining well above the VLM--VLA baseline throughout.}
    \label{fig:cosine_analysis}
\end{figure*}

\paragraph{Robocasa.} RoboCasa~\cite{robocasa365} is a large-scale simulation benchmark for household robot manipulation in photorealistic kitchen environments, comprising a set of atomic skills (e.g.\ pick-and-place, opening/closing doors and drawers, turning knobs, pressing buttons) together with longer-horizon composite tasks. We evaluate on three atomic kitchen tasks: \textbf{Open Fridge}, requiring the robot to open a refrigerator door; \textbf{Setup Cup}, requiring the robot to pick up a cup and place it in a target location; and \textbf{Navigate Kitchen}, requiring the robot to move within the kitchen environment to reach a target location before completing a manipulation sub-goal. Compared to LIBERO, RoboCasa tasks are evaluated across a larger number of randomised scene layouts and object instances, making the benchmark substantially more challenging: even strong VLA baselines such as $\pi_{0.5}$ average only around 39\% success rate across these tasks in our setting, reflecting the benchmark's greater diversity in scene configurations and the longer effective horizon of some tasks (e.g.\ Navigate Kitchen).
\begin{figure}[ht]
        \centering
    \begin{minipage}{0.48\linewidth}
        \centering
        \textbf{$\mathbf{Ours}_{\pi_{0.5}}$}\\[2pt]
        \includegraphics[width=\linewidth]{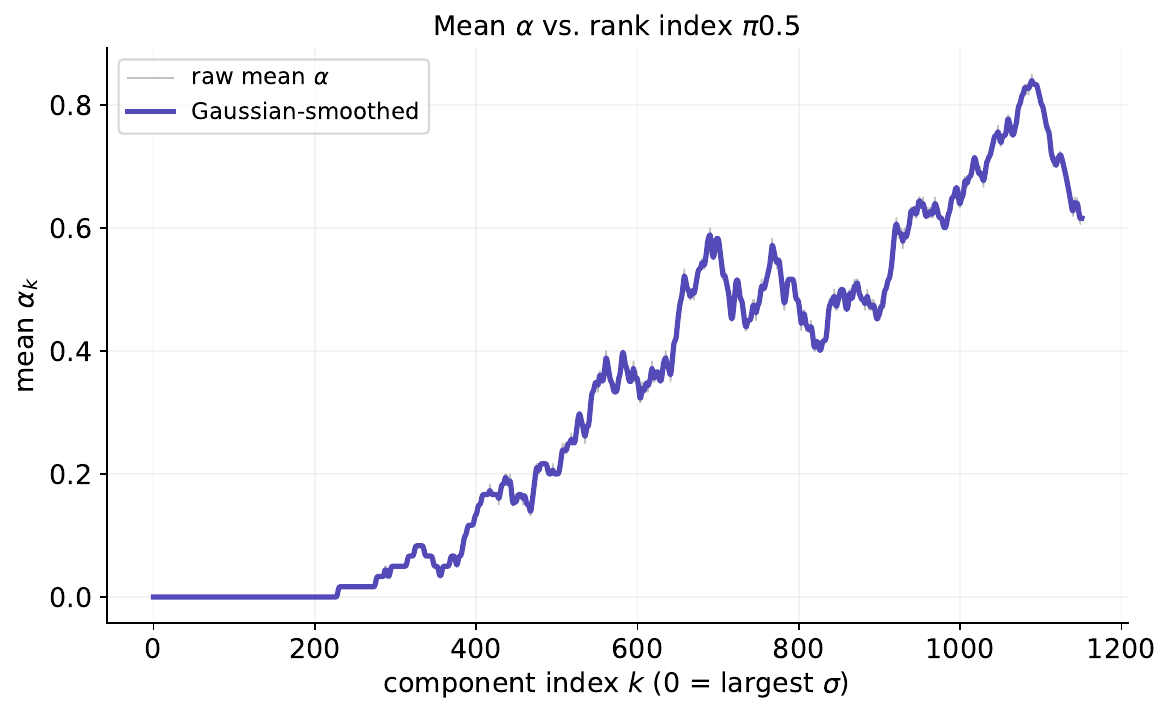}
    \end{minipage}
    \hfill
    \begin{minipage}{0.48\linewidth}
        \centering
        \textbf{$\mathbf{Ours}_{\pi_{0FAST}}$}\\[2pt]
        \includegraphics[width=\linewidth]{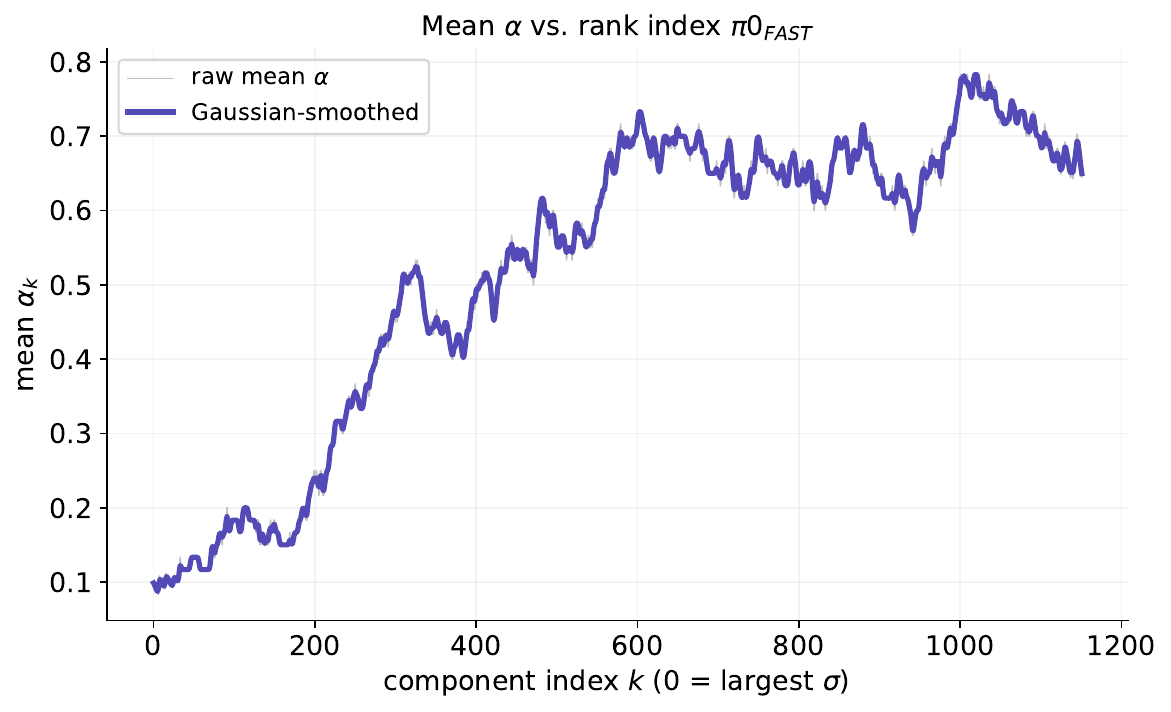}
    \end{minipage}
    \caption{\textbf{Mean $\boldsymbol{\alpha}$ vs.\ rank index.} \textit{(Top)}\textbf{ Ours}$_{\pi_{0.5}}$. \textit{(Bottom)}\textbf{ Ours}$_{\pi_{0Fast}}$. For each rank index $k$, we average $\alpha_k^{(l)}$ across all 602 layers. Both VLA variants show the same overall pattern: mean $\alpha$ is near 0 for the most dominant directions and rises monotonically toward the tail of the spectrum, confirming that this asymmetry is not specific to a single action decoding strategy.}
    \label{fig:alpha_vs_rank}
\end{figure}
\subsection{Additional Results on RoboCasa}

\label{app:robocasa}

Table~\ref{tab:robocasa} reports task success rate on three RoboCasa kitchen tasks — Open Fridge, Setup Cup, and Navigate Kitchen — complementing the LIBERO results presented in the main paper. The overall trends are consistent with those observed on LIBERO: the pure VLM achieves zero success rate everywhere, both pure VLAs establish strong baselines, and all fixed-coefficient merging methods (Model Soup, DARE, SLERP) collapse entirely to zero success rate across all tasks. Our merged models retain meaningful action performance on both VLA variants:\textbf{ Ours}$_{\pi_{0Fast}}$ reaches 56/61/78 (versus 78/68/89 for the pure $\pi_0$-FAST), and\textbf{ Ours}$_{\pi_{0.5}}$ reaches 65/49/21 (versus 71/58/25 for the pure $\pi_{0.5}$). The performance drop relative to the pure VLAs is more pronounced on RoboCasa than on LIBERO, reflecting the greater difficulty of this benchmark: even the pure VLAs average only around 50--78\% across tasks, and Navigate Kitchen proves particularly challenging for $\pi_{0.5}$ (25\% for the pure VLA, 21\% after merging). Despite this increased difficulty, our method remains the only merging approach that retains non-trivial action performance while simultaneously recovering the vision-language and embodied reasoning capabilities reported in the main paper.

\section{Analysis of the Merged Model}
\label{app:cosine_analysis}

\subsection{Models Similarity}
\label{app:similarity_analysis}

To better understand what the learned $\boldsymbol{\alpha}$ coefficients produce in practice, we analyse the per-layer alignment of the merged model with respect to both source models, for both VLA variants. For each layer $l$, we compute three cosine similarities: $\text{sim}(\theta_{\text{VLM}}^{(l)}, \theta_{\text{VLA}}^{(l)})$ — the baseline distance between the two source models — $\text{sim}(\theta_{\text{merged}}^{(l)}, \theta_{\text{VLM}}^{(l)})$, and $\text{sim}(\theta_{\text{merged}}^{(l)}, \theta_{\text{VLA}}^{(l)})$, where $\theta_{\text{merged}}$ denotes the merged model for the corresponding VLA variant. These are computed directly on the reconstructed merged weight matrices using the learned $\boldsymbol{\alpha}$.

Figure~\ref{fig:cosine_analysis} reports these three curves across all 602 layers for both VLA variants. Several observations emerge, most of which hold consistently across both panels. First, the baseline similarity between VLM and VLA (blue) is consistently the lowest in both cases, confirming that VLA fine-tuning significantly displaces the weights from their original VLM values; this displacement grows with layer depth, dropping from $\sim$0.80--0.85 in the earliest layers to $\sim$0.65--0.70 in the deepest ones, indicating that fine-tuning increasingly diverges from the VLM as depth increases, for both the autoregressive and flow matching VLA. Second, and most importantly, the merged model remains significantly closer to both source models than the VLM and VLA are to each other in both cases — orange and green curves lie well above the blue baseline throughout. This indicates that the merged model does not simply interpolate between two distant models but finds a position in weight space that is simultaneously close to both sources, with no significant drop in alignment relative to either, regardless of the underlying VLA architecture. Third, the merged model is consistently and substantially closer to the VLA than to the VLM across all layers in both panels (green above orange throughout), indicating that the learned $\boldsymbol{\alpha}$ coefficients globally preserve a large share of the VLA adaptation while still maintaining a clear margin of proximity to the VLM. The main difference between the two panels is the magnitude of this gap: for\textbf{ Ours}$_{\pi_{0.5}}$ (bottom), the green curve sits substantially higher ($\sim$0.97--0.98) and more tightly clustered than for\textbf{ Ours}$_{\pi_{0Fast}}$ (top, $\sim$0.88--0.93), while the orange curves are comparable in both cases ($\sim$0.83--0.93). This is consistent with the merged $\pi_{0.5}$ model retaining a stronger and more uniform alignment with the VLA across layers, in line with the higher action success rates observed for\textbf{ Ours}$_{\pi_{0.5}}$ in Table~1. Both the orange and blue curves exhibit a regular sawtooth pattern in both panels, consistent with the alternation between different weight types (attention and feed-forward) within each transformer block. Additionally, the orange curve partially recovers around layer index 350--400 before declining again toward the end of the network in both cases. This structured behaviour emerges directly from the optimisation of $\boldsymbol{\alpha}$, confirming that the learned coefficients successfully navigate the trade-off between both source models rather than collapsing toward either one, irrespective of the VLA's action decoding strategy.

\subsection{Merging coefficients analysis}
\label{app:alpha_rank}

To examine how the learned coefficients are distributed across the singular spectrum, we average $\alpha_k^{(l)}$ across all layers at each rank index $k$ (where $k=0$ corresponds to the largest singular value $\sigma_0^{(l)}$ within each layer), for both VLA variants.

Figure~\ref{fig:alpha_vs_rank} shows the same overall pattern for both VLA variants: mean $\alpha$ is near 0 for the most dominant directions of the task vector and rises monotonically toward the tail of the spectrum. For\textbf{ Ours}$_{\pi_{0.5}}$ (top), mean $\alpha$ is exactly 0 for $k<200$ and rises to a peak around 0.8 near $k\approx1050$ before declining slightly toward the tail. For\textbf{ Ours}$_{\pi_{0Fast}}$ (bottom), the trend is similar, with mean $\alpha$ starting close to, though not exactly at, 0 ($\sim$0.1 at $k=0$) and plateauing around 0.65--0.7 from $k\approx600$ onward. The consistency of this monotonic pattern across both VLA architectures indicates that the low-rank/high-rank asymmetry exists between language and action recovery. This pattern directly complements the rank ablation of Section~\ref{sec:ablation}: when the model is restricted to a low rank, only the dominant directions are available, leaving the optimisation with little room to inject VLA-relevant adaptation, which is consistent with action prediction degrading sharply at low rank while language recovers comparatively quickly, for both VLA variants. Conversely, the gradual rise of mean $\alpha$ toward the tail of the spectrum in both panels shows that the directions necessary for action prediction are spread broadly across the weaker components, consistent with the diffuse nature of the VLA adaptation observed throughout this work.

\subsection{Additional Qualitative Examples}
\label{app:qualitative}

\begin{figure*}[htbp]
    \centering
    \includegraphics[width=\linewidth]{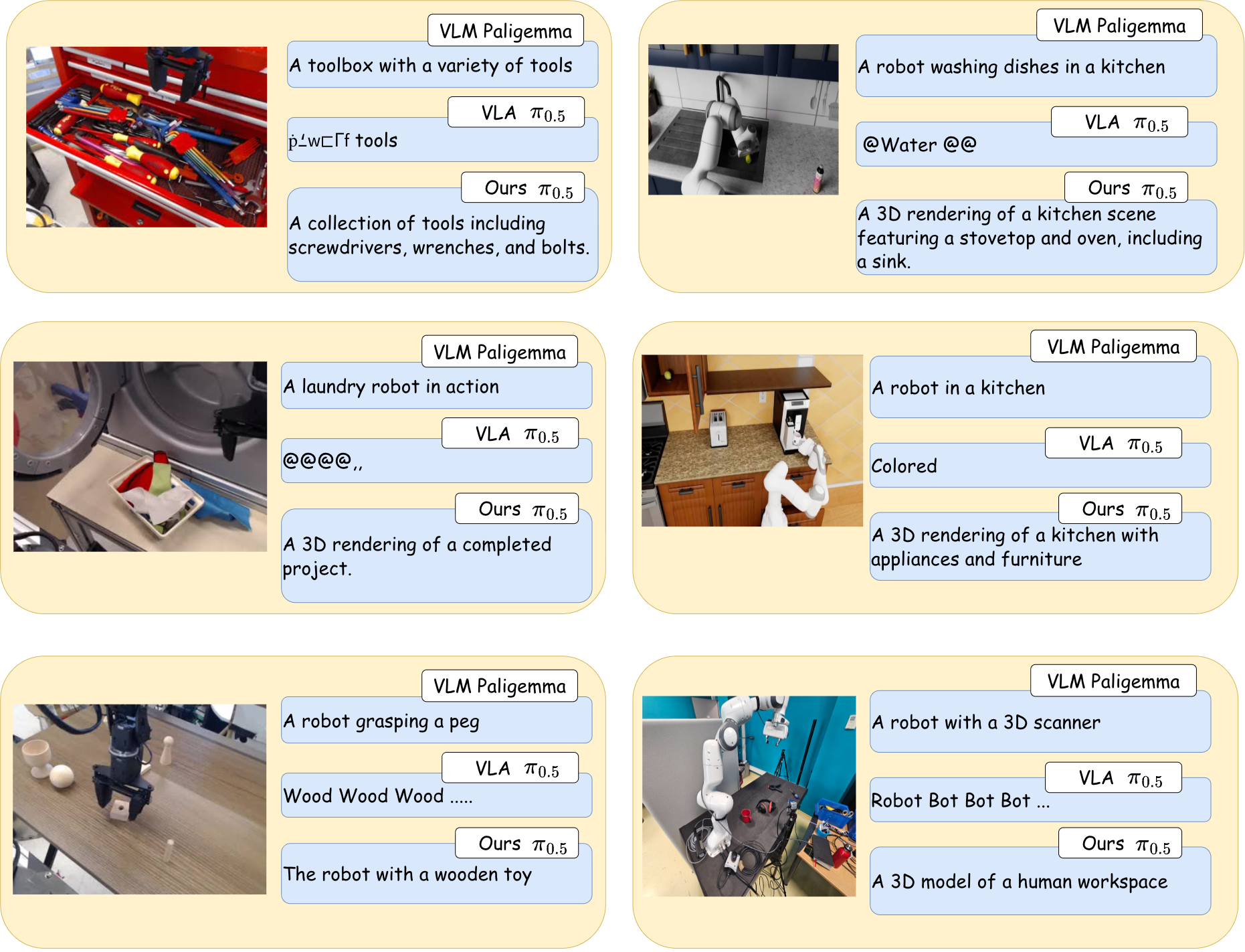}
    \caption{\textbf{Qualitative captioning examples.} Given the same visual observation, the VLM produces coherent scene descriptions, the VLA ($\pi_{0.5}$) generates fragmented or incomplete text due to catastrophic forgetting, and \textbf{Ours}$_{\pi_{0.5}}$ merged model recovers meaningful captions while retaining action capabilities.}
    \label{fig:app qualitative}
\end{figure*}

Figure~\ref{fig:app qualitative} shows six additional captioning examples spanning diverse scenes, from tabletop manipulation to kitchen environments. Across all examples, the pure VLA output degrades into either a single repeated keyword (e.g.\ ``Wood Wood Wood ....'', ``Robot Bot Bot ...''), a short fragment stripped of grammatical structure (e.g.\ ``Colored'', ``@Water @@''), or garbled non-linguistic tokens (e.g.\ ``\.{p}\d{}w\r{}$\Gamma$ f tools''), consistent with the catastrophic forgetting of language modelling capabilities discussed in Section~\ref{sec:vlm}. In contrast, our merged model consistently recovers full, grammatically coherent sentences that correctly identify the objects and setting present in the scene (e.g.\ ``A collection of tools including screwdrivers, wrenches, and bolts.'', ``A 3D rendering of a kitchen scene featuring a stovetop and oven, including a sink.''). We note that our merged model's captions occasionally diverge stylistically from the pure VLM's — favouring phrases such as ``3D rendering of...'' — but remain semantically accurate and fluent, indicating that language coherence is recovered even when the exact phrasing differs from the original VLM.
These qualitative observations are consistent with the quantitative captioning scores reported in Table~\ref{tab:vlm}, where the pure VLA collapses on ROUGE-L, METEOR and BERTScore while our merged model recovers most of the pure VLM's performance.
\end{document}